\pdfoutput=1

\documentclass[10pt, twocolumn, letterpaper]{article}

\usepackage{cvpr}              

\makeatletter
\@namedef{ver@everyshi.sty}{}
\makeatother
\usepackage{tikz}

\usepackage{graphicx}
\usepackage{caption}
\usepackage{amsmath}
\usepackage{amssymb}
\usepackage{booktabs}
\usepackage{multirow}
\usepackage{url}            
\usepackage{amsfonts}       
\usepackage{float}
\usepackage{microtype}      
\usepackage{nicefrac}       
\usepackage{xcolor}         
\usepackage{mathrsfs}
\usepackage{comment}

\usepackage[accsupp]{axessibility}  

\def\authorBlock{
    Xiangyu Zhu${}^{12*}$ \quad
    Dong Du${}^{1}$\thanks{Xiangyu Zhu and Dong Du contribute equally.} \quad
    Weikai Chen${}^{3}$ \quad
    Zhiyou Zhao${}^{1}$ \quad \\
    Yinyu Nie${}^{4}$ \quad
    Xiaoguang Han${}^{12}$\thanks{Corresponding author's email: hanxiaoguang@cuhk.edu.cn} \quad
    \\
    \\
    ${}^{1}$SSE, CUHKSZ  \quad    ${}^{2}$FNii, CUHKSZ \quad ${}^{3}$Tencent America \quad ${}^{4}$Technical University of Munich
}

\usepackage[pagebackref,breaklinks,colorlinks]{hyperref}

\usepackage[capitalize]{cleveref}
\crefname{section}{Sec.}{Secs.}
\Crefname{section}{Section}{Sections}
\Crefname{table}{Table}{Tables}
\crefname{table}{Tab.}{Tabs.}

\definecolor{DeltaColor}{rgb}{0.039,0.73,0.71}
\definecolor{SetaColor}{rgb}{0.867, 0.0235, 0.376}
\definecolor{SigmaColor}{rgb}{0.98,0.45,0.0}
\definecolor{RedColor}{rgb}{0.8,0,0}
\definecolor{AlphaColor}{rgb}{1.0, 0.4, 0.8}
\definecolor{BetaColor}{rgb}{0.8,0,0.8}
\definecolor{GammaColor}{rgb}{0.0,0,0.7}
\definecolor{EpsilonColor}{rgb}{0.353,0.725,0.906}
\definecolor{TauColor}{rgb}{0.423,0.235,0.192}

\newcommand{\ReviewOne}[1]{{\color{RedColor}TRnS}}

\newcommand{\modelName}{NerVE}

\newcommand{\symbcubeocc}{$o$}
\newcommand{\symbcubeori}{$e$}
\newcommand{\symbcubepos}{$p$}
\newcommand{\maskcubeocc}{$\mathcal{M}_o$}
\newcommand{\maskcubeori}{$\mathcal{M}_e$}
\newcommand{\maskcubepos}{$\mathcal{M}_p$}

\newcommand{\RecallCube}{$R_o$}
\newcommand{\PrecisionCube}{$P_o$}
\newcommand{\CorrectFace}{$C_e$}
\newcommand{\DistancePoint}{$D_p$}

\newcommand{\DegreeOne}{$\mbox{1-d }$}

\begin{document}

\title{\modelName{}: Neural Volumetric Edges for Parametric \\ Curve Extraction from Point Cloud}

\author{\authorBlock}

\maketitle

\begin{abstract}
Extracting parametric edge curves from point clouds is a fundamental problem in 3D vision and geometry processing.
Existing approaches mainly rely on keypoint detection, a challenging procedure that tends to generate noisy output, making the subsequent edge extraction error-prone.
To address this issue, we propose to directly detect structured edges to circumvent the limitations of the previous point-wise methods.
We achieve this goal by presenting \modelName{}, a novel neural volumetric edge representation that can be easily learned through a volumetric learning framework.
\modelName{} can be seamlessly converted to a versatile piece-wise linear (PWL) curve representation, enabling a unified strategy for learning all types of free-form curves. Furthermore, as \modelName{} encodes rich structural information, we show that edge extraction based on \modelName{} can be reduced to a simple graph search problem.
After converting \modelName{} to the PWL representation, parametric curves can be obtained via off-the-shelf spline fitting algorithms.
We evaluate our method on the challenging ABC dataset~\cite{koch2019abc}.
We show that a simple network based on \modelName{} can already outperform the previous state-of-the-art methods by a great margin. Project page: \href{https://dongdu3.github.io/projects/2023/NerVE/}{https://dongdu3.github.io/projects/2023/NerVE/.}
\end{abstract}


\section{Introduction}
\label{sec:intro}



The advances of 3D scanning techniques have enabled us to digitize and reconstruct the physical world, benefiting a wide range of applications including 3D modeling, industrial design, robotic vision, \textit{etc}. 
However, point clouds, the raw output of a 3D scanner, are typically noisy, unstructured, and can exhibit strong sampling bias.
Hence, extracting structured features, such as the feature edges, from an unordered point cloud is a vital geometry processing task.
Sharp geometric edges can be used as an abstraction of a complex 3D shape, facilitating downstream tasks including surface reconstruction, normal estimation, and shape classification.
Previous state-of-the-art methods mainly resort to a \emph{keypoint fitting} strategy to extract parametric edge curves from a point cloud.
Specifically, they first detect a sparse set of keypoints, such as the endpoints or points on sharp edges, and then group these points into individual sets according to predefined topologies.
Finally, each point set is converted into a parametric curve using spline fitting.

\begin{figure}[!t]
	\centering
	\includegraphics[width=1.\linewidth]{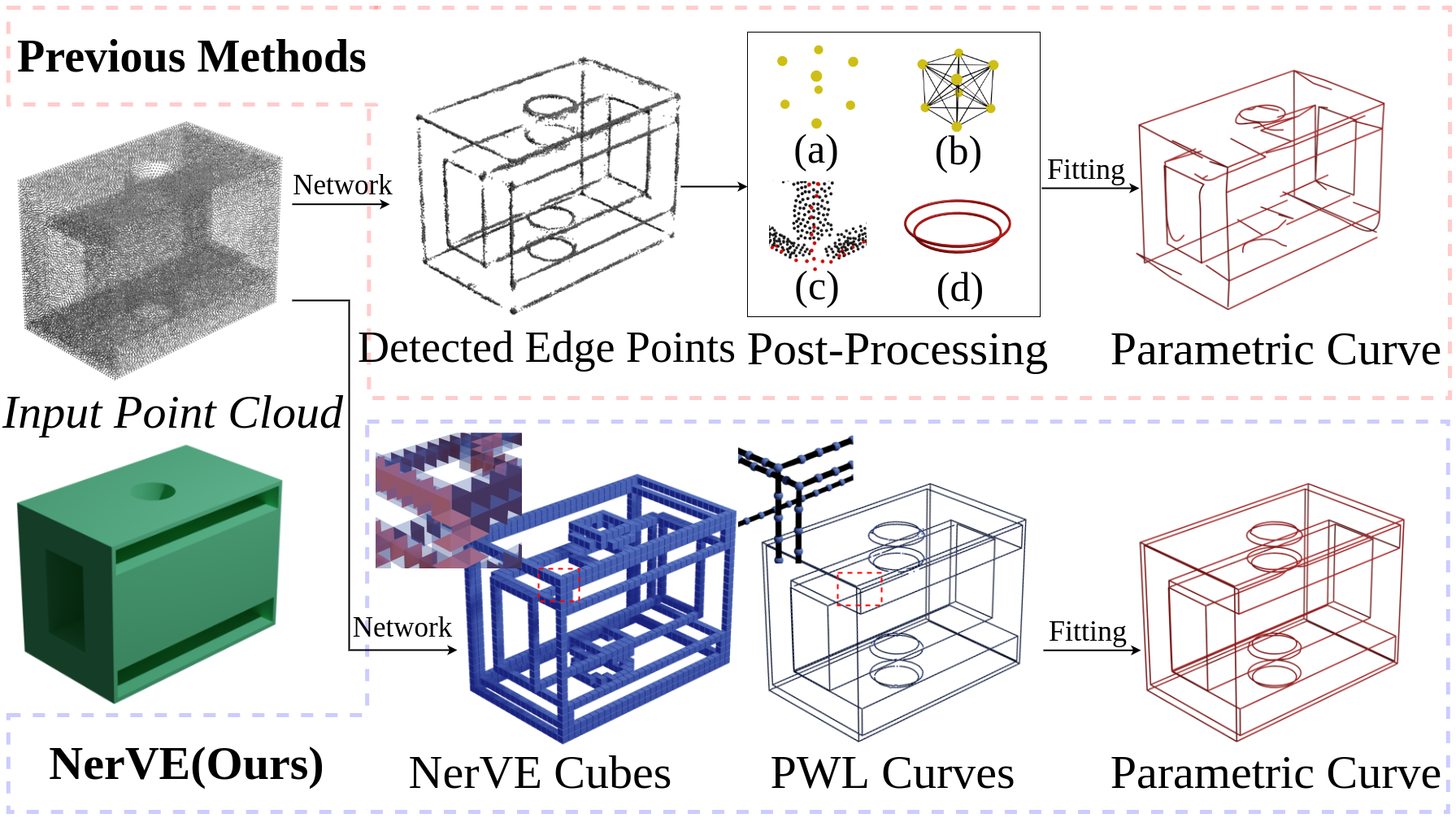}
	\caption{We present NerVE, a neural volumetric edge representation for parametric curve extraction from point clouds. Our method can predict structured NerVE instead of unstructured edge points in previous methods, and directly convert NerVE into piece-wise linear (PWL) curves, reducing the error-prone post-processing of previous methods into simple graph search on PWL curves. Post-processing includes (a) Endpoint detection;(b) Graph structure analysis;(c) Points thinning;(d) Special treatment for closed curves.}
	\label{fig:teaser}
	\vspace{-5mm}
\end{figure}

Recent approaches have strove to improve the accuracy of edge point detection by using hand-crafted features~\cite{merigot2010voronoi} or deep neural networks~\cite{yu2018ec,wang2020pie,bazazian2021edc,matveev2022def}. 
Despite the impressive progress that has been made, existing works still have the following limitations.
1) The widely adopted \emph{point-wise} classification approaches tend to generate noisy estimations -- the predicted edge points typically contain a spurious set of candidate points (see Fig.~\ref{fig:comp_edgepoints_vis}), which requires further processing for keypoint cleaning and increases the risk of false/missing connections.
2) The grouping procedure highly relies on the accuracy of endpoint detection. 
However, it remains difficult to accurately locate endpoints especially when the normal of its surrounding points change smoothly.
3) They require tedious treatments to cope with different curve topologies, including curve type estimation, topology-dependent artifact points removal and curve connection, \emph{etc}.

Our key observation is that the above issues can be resolved if we can \emph{directly predict structured edges} in the form of \emph{piece-wise linear} (PWL) curves from the input point cloud. 
This bypasses the problematic keypoint detection and avoids the error-prone edge extraction in the curve fitting stage.
In addition, PWL curve is a general representation of free-form curves, removing the need of curve topology estimation and the laborious curve fitting and post-processing dependent on curve category.
Furthermore, PWL curves can be easily converted to parametric curves using the off-the-shelf solutions.
However, unlike its parametric counterpart, PWL curves are notoriously difficult to predict due to its large degrees of freedom.  

Towards this end, we propose \emph{\modelName{}}, a novel neural edge representation in a volumetric fashion. 
As shown in Fig.~\ref{fig:cube_definition},  \modelName{} represents 3D structured edges using a regular grid of volumetric cubes -- each cube encodes rich structural information including 1) one binary indicator of edge occupancy, 2) edge orientations (if any), and 3) one edge point position (if any).
Thereby, \modelName{} can be readily converted to PWL curves by connecting the edge points enclosed by \modelName{} cubes according to the encoded point connectivity.
The introduction of \modelName{} brings several advantages. 
First, the generated \modelName{} cubes are structured by itself, which greatly simplifies the process of curve extraction.
Second, it is fully compatible with the PWL curve representation, and hence, can deal with all types of curves in a unified manner.
Third, \modelName{} cubes can be viewed as a coarse representation of the point cloud. Predicting the occupancy of a volumetric cube is easier and more robust than point-wise classification. 
Therefore, we are less likely to suffer from the issue of missing curves (see our results in Fig.~\ref{fig:comp_paramcurve}). 
Lastly, inferring \modelName{} can be formulated as a voxel-wise classification and regression problem, where the well-developed 3D convolutional networks can be directly employed.

We further propose a volumetric learning framework to predict \modelName{} from the input point cloud. 
We first encode the point features into a volumetric feature grid with the same resolution of the output.
Then, a multi-head decoder is used to predict the attributes of a \modelName{} cube from its corresponding feature cell.
After converting the \modelName{} cubes into PWL curves, a specially-tailored post-processing procedure is proposed to correct potential topology errors in the resulting curves. 
Finally, the parametric curves can be obtained via a straightforward spline fitting algorithm. 

We evaluate our method on the ABC dataset~\cite{koch2019abc}, a large-scale collection of computer-aided design (CAD) models with challenging topology variations.
In particular, we compare with the state-of-the-art approaches on two different tasks: edge estimation and parametric curve extraction.
Experimental results show that by leveraging the proposed \modelName{} representation, our method can faithfully extract complete and accurate edges and parametric curves from intricate CAD models, outperforming the other methods both qualitatively and quantitatively.

We summarize our contributions as follows:
\begin{itemize}
   \item We propose \emph{NerVE}, a learnable neural volumetric edge representation that supports direct estimation of structured 3D edges, seamless conversion with general PWL curves, and compatibility with latest volumetric learning framework. 

   \item A pipeline for parametric curve extraction from point cloud that consists of a learning-based framework for faithful  \modelName{} cubes estimation and a post-processing module for curve topology correction. 
   
   \item We set a new state-of-the-art on the ABC dataset  in the task of parametric curve extraction from point cloud. 
\end{itemize}

\section{Related Works}
\label{sec:related}

\noindent \textbf{Edge Feature Detection.}
Traditional edge feature detection from a point cloud is based mainly on local geometric features, such as the eigenstructure of the covariance matrix~\cite{gumhold2001feature,pauly2003multi,bazazian2015fast,xia2017fast}, normals~\cite{fleishman2005robust,demarsin2006detection,demarsin2007detection,weber2010sharp}, curvatures~\cite{lin2015line,hackel2016contour}, or other statistical metrics~\cite{weber2011methods,nie2016extracting}.
To improve the robustness of edge detection for noisy point clouds, Daniels et al.~\cite{daniels2007robust} use a multistep refinement method with robust moving least squares to fit the surface to potential features. Ni et al.~\cite{ni2016edge} combine RANSAC and the angular map metric to detect edges. VCM~\cite{merigot2010voronoi} is presented by measuring Voronoi covariance and applying the Monte-Carlo algorithm to compute feature boundaries, which is widely used in geometry processing. With the bloom of deep learning, PIE-Net~\cite{wang2020pie}, EDC-Net~\cite{bazazian2021edc}, and PCEDNet~\cite{himeur2021pcednet} are proposed to formulate the edge detection as a classification task and utilize neural networks to learn it. On the other hand, EC-Net~\cite{yu2018ec} reformulates it as a regression problem, then learns residual point coordinates and point-to-edge distances to identify edge points. All of these learning-based methods significantly improve the accuracy of edge detection. In this paper, we also utilize advanced neural networks but avoid explicit edge detection. We propose a novel \modelName{} representation to directly learn the positions and topology of edge points. 

\begin{figure*}[!t]
	\centering
	\includegraphics[width=1.\linewidth]{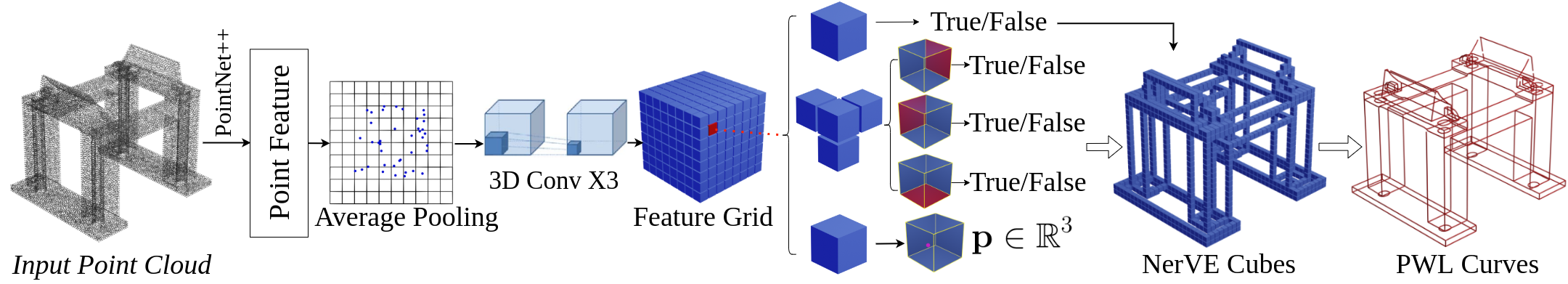}
	\caption{Overview of our proposed network for learning NerVE. Given a point cloud, we first utilize a simplified PointNet++~\cite{qi2017pointnet++} module and a 3D CNN module to obtain the feature grid (has the same resolution of our \modelName{} output). Three individual decoders are applied to process cube features in the grid to predict the corresponding three attributes of NerVE, i.e., edge occupancy, edge orientation, and edge point position. By converting the NerVE cubes into PWL curves, we can obtain the parametric curves of the shape with a post-processing.}
	\label{fig:pipeline}
	\vspace{-5mm}
\end{figure*}


\vspace{1mm}
\noindent \textbf{Parametric Curve Extraction.}
Parametric curves have been widely used in CAD modeling to design complex 3D shapes with sharp geometries. However, extracting parametric curves from a point cloud is challenging due to the various types and complex connections of curves. Pioneering work~\cite{gumhold2001feature} attempts to detect feature points and fits them with splines to recover feature lines. Recent deep learning methods~\cite{wang2020pie,matveev20213d,liu2021pc2wf,tan2022coarse,matveev2022def,guo2022complexgen} leverage a keypoint fitting strategy that first detects a sparse set of keypoints (e.g., corner points and edge points), then groups and connects them, followed by a determination of the target parametric curve type and performs curve fitting. Specifically, PIE-Net~\cite{wang2020pie} utilizes a PointNet++~\cite{qi2017pointnet++}-like network to extract point features for the classification of edge points, corners, and others, then generates curve proposals for parametric curve extraction. ComplexGen~\cite{guo2022complexgen} formulates the prediction of validness and primitive types as classification tasks, and recovers corners, curves, and patches simultaneously along with their mutual topology constraints. In contrast, DEF~\cite{matveev2022def} regresses a continuous distance field to represent the distance from the input points to the closest feature lines, and then extracts parametric feature curves from the inferred field. Other works~\cite{matveev20213d,liu2021pc2wf,tan2022coarse} simplify the curves into lines only, and focus on 3D wireframe reconstruction. However, complicated corner detection and connection estimation make these keypoint fitting methods prone to produce artifacts (e.g., missing curves). To this end, we propose \modelName{}, a novel neural-based edge representation that supports the prediction of structured edges in the form of PWL curves, making the following parametric curve extraction easy and efficient.

\noindent \textbf{Neural Representation Learning.}
Many well-known 3D representations, such as voxels~\cite{choy20163d}, point clouds~\cite{fan2017point}, meshes~\cite{wang2018pixel2mesh,groueix2018papier}, occupancy fields~\cite{mescheder2019occupancy}, and signed distance functions~\cite{park2019deepsdf,chen2019learning}, have been introduced into deep learning to resolve the problem of 3D reconstruction and achieve impressive results. Liao et al.~\cite{liao2018deep} propose a differentiable learning architecture to represent the classical Marching Cubes~\cite{lorensen1987marching} algorithm for shape reconstruction. Chen et al.~\cite{chen2021neural,chen2022neural} further extend the algorithms of Marching Cubes~\cite{lorensen1987marching} and Dual Contouring~\cite{ju2002dual} with data-driven approaches. Specifically, they implicitly represent triangle meshes in compact per-cube parameterizations that are compatible with neural learning. With the learned implicit field, a high-quality triangle mesh with sharp features can be directly extracted. Inspired by the tendency, we introduce the traditional PWL curve representation into deep learning for parametric curve estimation. To make it easy to learn PWL curves, we propose \modelName{} to parameterize PWL curves in a uniform volumetric field and apply advanced neural networks for learning. 

\begin{figure}[h]
	\centering
	\includegraphics[width=0.94\linewidth]{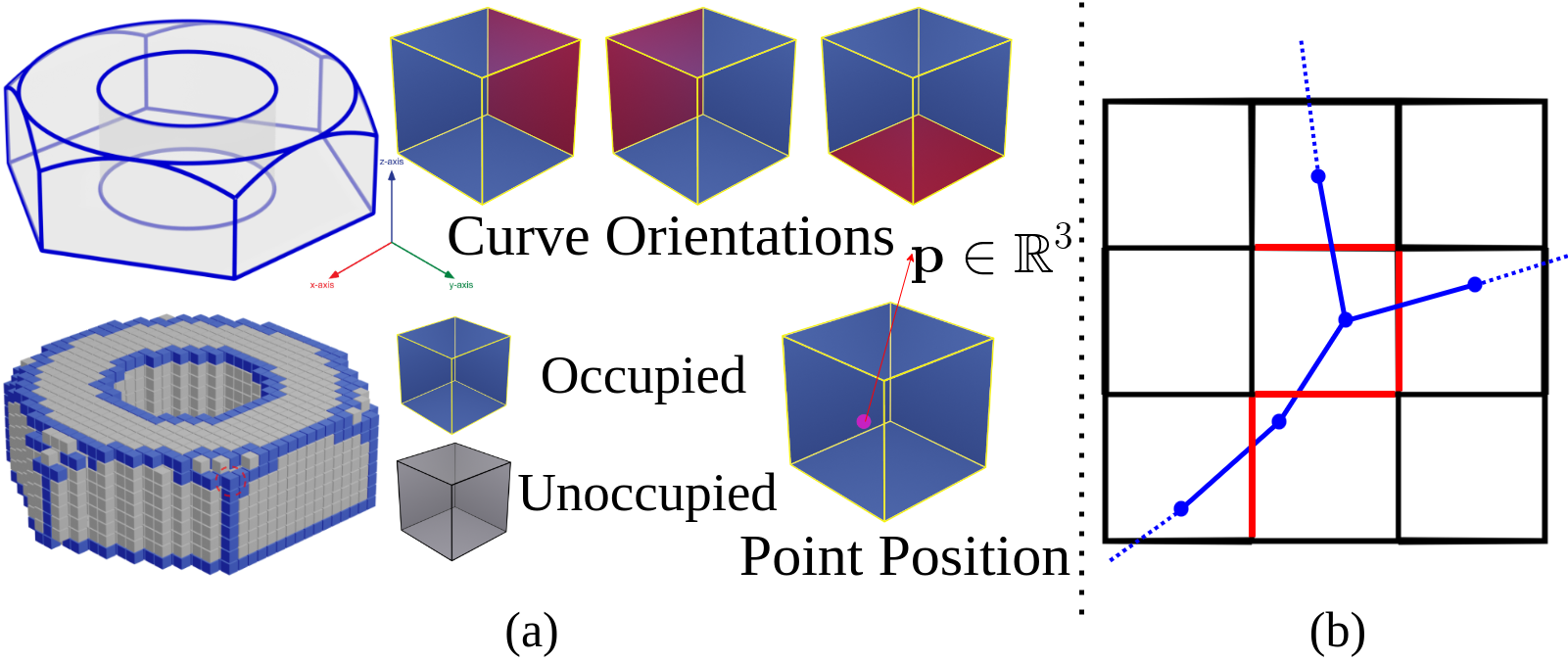}
	\caption{(a) Definition of the three attributes in NerVE. (b) Illustration of PWL curve extraction from NerVE. A red edge is corresponding to the red face in (a), showing there exists at least one curve passing through the face, i.e., the edge orientation is true. }
	\label{fig:cube_definition}
	\vspace{-5mm}
\end{figure}
\section{Method}
\label{sec:method}

\noindent \textbf{Overview.} Different from traditional keypoint fitting workflows~\cite{yu2018ec,wang2020pie,bazazian2021edc,matveev2022def}, we present a new paradigm to generate accurate parametric curves using a neural volumetric edge representation, named \emph{NerVE}. First, we introduce the definition of 
NerVE and then propose a dedicated network to learn NerVE that supports direct estimation of structured 3D edges.
These edges compose PWL curves that can not only provide precise edge positions but also contain valuable topology information, which eases the extraction of parametric curves. For the inferred PWL curves, we further utilize a simple post-processing procedure to correct their topology errors. Finally, parametric curves can be obtained using a straightforward graph search and spline fitting algorithm.

\subsection{NerVE Representation}

Learning 3D parametric curves from a point cloud is a challenging task due to the complexity of both curve types and connections. 
Existing works mainly resort to a keypoint fitting strategy and tend to generate noisy or error-prone results. In this work, we propose a unified edge representation, \modelName{}, that is compatible with different types of curves and is also feasible to be inferred with the latest volumetric learning techniques.

\subsubsection{Volumetric Edge Definition}
\label{sec:ve_def}

We follow the voxel representation of a 3D shape~\cite{choy20163d} to discretize continuous 3D curves into a volumetric grid. As shown in Fig.~\ref{fig:cube_definition}, each cube in NerVE contains three attributes: 1) edge occupancy \symbcubeocc{}. $o\in\{0, 1\}$, which is a binary value to define whether the cube contains edges; 2) edge orientations \symbcubeori{}. $e_i\in\{0, 1\}, i=1,2,3$, which are 3 binary values that represent whether the cube should connect with its adjacent cubes to construct edge pieces. Note that a cube has 6 faces and shares with its neighbors, thus we define only three statuses on the left, bottom, and back faces in a cube; 3) edge point position \symbcubepos{}, which defines an edge point in the cube. With the three defined attributes, we can discretize continuous curves in a unified and regular volumetric representation, which can be learned directly via neural networks.

For NerVE, the higher resolution could keep more curve information but also increase the calculation consumption for network training and curve extraction. In our method, we use the resolution of $32^3$, which we found it works well in our experiments. We provide the ablation of different NerVE resolutions in Sec.~\ref{exp:ablation}. Regarding the determination of ground-truth  \symbcubepos{}, we generally choose the midpoint of the truncated curve inside a cube, which is natural and works well. For a special case where multiple curves appears in a cube, we calculate the average midpoint position or pick the endpoint (if any) as \symbcubepos{}. We also provide an ablation study about the selection of \symbcubepos{} in Sec.~\ref{exp:ablation}.


\subsubsection{PWL Curve Extraction}

Given our volumetric edge representation (i.e., NerVE), the underlying curves can be easily extracted in the form of PWL curves. Specifically, we pick the cube points $\{p\}$ when their occupancy statuses are true, and connect them with their neighbor points if the edge orientation statuses are true. Hence, a PWL curve is obtained, as illustrated in Fig.~\ref{fig:pipeline}. 

\subsection{NerVE Learning}

Given a point cloud $\mathcal{X}$ in the 3D space, we adopt a dense and simplified PointNet++~\cite{qi2017pointnet++} module and a simplified 3D CNN module~\cite{choy20163d} to extract efficient local features for the unorganized point cloud, then use an MLP-based module to generate our NerVE, denoted as $\{y_{(i,j,k)}\}$, where $y=(o,e,p)$ is the volumetric cube attributes at the position of $(i,j,k)$. We can formulate it as learning a function $\mathcal{F_\theta}$,
\begin{equation}
    \mathcal{F}_\theta: \mathcal{X} \longrightarrow \{y_{(i,j,k)}\}, \forall (i,j,k),
\end{equation}
where $\theta$ denotes the parameters in a neural network.

\subsubsection{Network Architecture}

Our network adopts an encoder-decoder paradigm to individually learn each attribute ($o, e, p$) in NerVE cubes (see Fig.~\ref{fig:pipeline}). Note that the encoder is comprised of point convolutions (1-dimensional convolution) and volume convolutions (3D CNN). We use the average pooling to connect these two kinds of convolutions, 
providing a feature grid for the following learning. The decoder is based on MLP layers.

\vspace{1mm}
\noindent \textbf{Encoder.} Our encoder is a dense and simplified PointNet++~\cite{qi2017pointnet++} combined with a 3D CNN module. Specifically, for each input point $x\in\mathcal{X}$, we first calculate its $k$-nearest neighbors ($k=8$ in our experiments, following~\cite{chen2022neural}), and then apply 4 MLP layers, each layer followed by a leaky ReLU activation, to extract 128-dimension point features. Then we utilize average pooling to fuse the features of points if they fall into the same cubes. Hence, we encode the point features into cubes. For these cubes without points inside, we initialize their features with zeros for the following 3D CNN learning. The 3D CNN has 3 3D-convolution layers, each having a kernel size of 3, stride length of 1, and padding size of 1. A leaky ReLU activation is appended after each convolution layer. 

\vspace{1mm}
\noindent \textbf{Decoder.} We apply three decoders to learn the three cube attributes, respectively. Each decoder is comprised of 5 MLP layers with a leaky ReLU activation except the final one. Using the 128-dimension volumetric features from our encoder as input, we decode each cube feature to predict an occupancy \symbcubeocc{} and a point \symbcubepos{}. For the learning of edge orientation, we concatenate a cube feature with its three adjacent ones before decoding, as shown in Fig.~\ref{fig:pipeline}. Furthermore, we use a sigmoid function for the outputs of \symbcubeocc{} and \symbcubeori{}. 

\subsubsection{Training}
\label{sec:training}

Compared to the 3D shape volume, the edge volume representation is much more sparse. To address the data imbalance problem in training and reduce the calculation consumption, we learn each cube attribute with specific masks, denoted \maskcubeocc{}, \maskcubeori{}, \maskcubepos{}. Specifically, given a point cloud, 
we pick all the volumetric surface cubes (\maskcubeocc{}) to train their edge occupancy $\{o\}$. The edge orientations $\{e\}$ and edge points $\{p\}$ are learned from those occupied edge cubes. Thus, we use the ground-truth edge cubes as masks \maskcubeori{}, \maskcubepos{} to constrain that the edge orientations and edge point positions are learned in edge cubes during training.
During inference, we use the inferred edge occupancies $\{o\}$ as the mask to predict their edge orientations $\{e\}$ and edge points $\{p\}$. 


To train our network, we choose the binary cross entropy (BCE) loss for the learning of edge occupancy \symbcubeocc{} and edge orientation \symbcubeori{}, and adopt the $L_1$ loss for the regression of edge point \symbcubepos{}, i.e., 

\begin{align}
    L_o &= 1/|\mathcal{M}_o| \sum_{o \in \mathcal{M}_o} \text{BCE}(o, o_{gt}),
\end{align}
\begin{align}
    L_e &= 1/|\mathcal{M}_e| \sum_{e \in \mathcal{M}_e} \text{BCE}(e, e_{gt}), \\
    L_p &= 1/|\mathcal{M}_p| \sum_{p \in \mathcal{M}_p} \left \| p - p_{gt} \right \|_1,
\end{align}
where $o_{gt}, e_{gt}, p_{gt}$ are the ground-truth cube attributes in the corresponding masked cubes.


\begin{figure}[!t]
	\centering
	\includegraphics[width=0.96\linewidth]{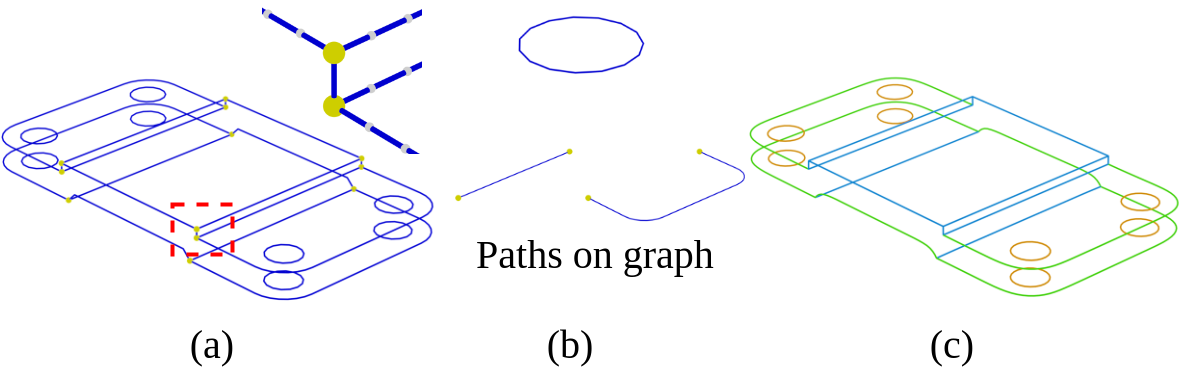}
	\caption{Parametric curve extraction via graph search. (a) The points with degree$>2$ are regarded as endpoints. (b) Finding a path between two endpoints or a closed path by searching on the PWL curve graph. (c) Final parametric curves after fitting the path.
	}
	\label{fig:pwl2cad}
	\vspace{-9mm}
\end{figure}

\subsection{Parametric Curve Extraction}

Most existing methods for parametric curve estimation struggle with the endpoint estimation \& connection, and the curve type recognition (e.g., line, circle, B-spline) from a point cloud in the learning stage. In contrast, our representation of \modelName{} is unified for learning various types of curves. \modelName{} can directly predict structured edges in the form of PWL curves, establishing a topology graph of edge points, as shown in Fig.~\ref{fig:pwl2cad}. The parametric curve extraction is then reduced to a simple graph-search problem. In this paper, we focus on curve extraction of CAD models.

For CAD models which are known as boundary representation (B-Rep), there are two constraints for parametric curves: 1) one open curve has only two endpoints, and two open curves can only be intersected at one endpoint; 2) the closed curves have zero endpoints~\cite{hofmann1989geometric,stroud2006boundary}. Given the PWL curves estimated from the inferred NerVE, we first choose the edge points with degree$>$2~\cite{weiler1986topological,stroud2006boundary} as endpoints. Here, the point degree is the number of edges that connect with the point in the PWL curves. Then we start from an endpoint and search along the PWL curve to obtain another endpoint. We thus fit with these two endpoints and all edge points on the path using an off-the-shelf spline fitting library. For the closed curve extraction, we just pick an edge point with a degree of 2 and search along the PLW curve graph until it ends up, and then fit with all points on the path. Our PWL curve graph greatly simplifies parametric curve extraction compared to previous methods.

However, it is inevitable to produce artifacts for a learning-based method, reducing the accuracy of curve extraction. To this end, we propose a straightforward post-processing method to refine the topology of PWL curves extracted from NerVE. Specifically, we connect two points with degree 1 if the distance between these two points is smaller than a given threshold and their tangent vectors are consistent. We also remove superfluous curve segments if their lengths are very short. More details are provided in the supplemental.
\section{Experiment}
\label{sec:exp}

In this section, we evaluate the effectiveness of our method with learning-based methods~\cite{yu2018ec,wang2020pie,matveev2022def} and a traditional method~\cite{merigot2010voronoi}. Then we test the robustness by adding noise or changing the sampling density of a point cloud, and provide an ablation study to explore the effects of certain choices in our method.

\begin{figure*}[!t]
	\centering
	\includegraphics[width=1.\linewidth]{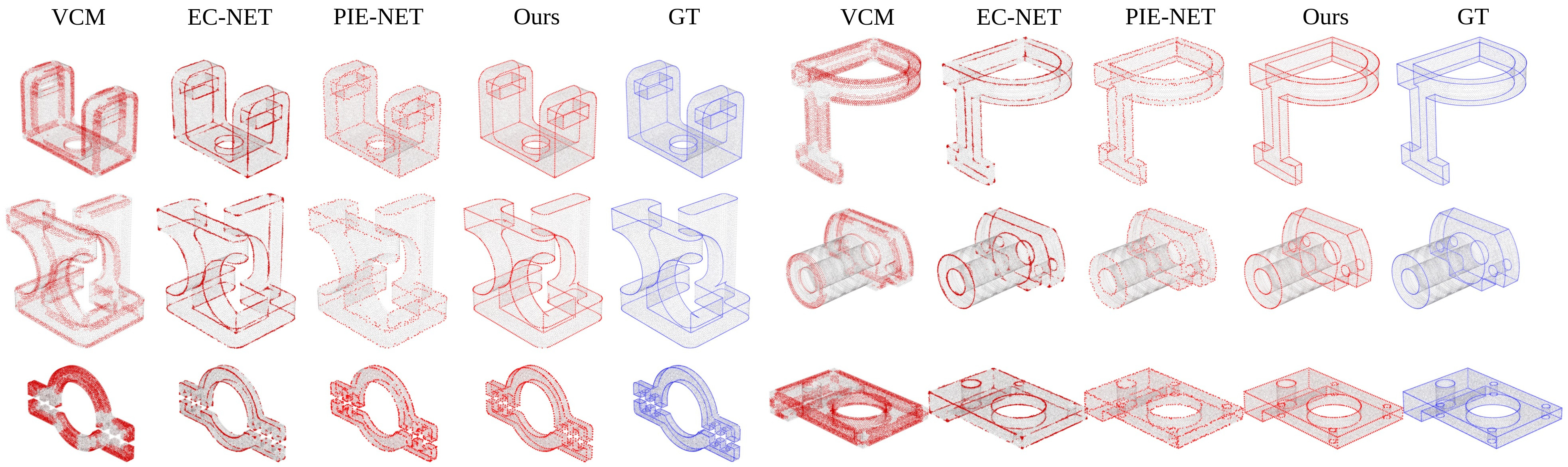}
	\caption{Qualitative comparisons on the edge point estimation with VCM~\cite{merigot2010voronoi}, EC-NET~\cite{yu2018ec} and PIE-NET~\cite{wang2020pie}.}
	\label{fig:comp_edgepoints_vis}
	\vspace{-1mm}
\end{figure*}

\begin{figure*}[!t]
	\centering
	\includegraphics[width=1.\linewidth]{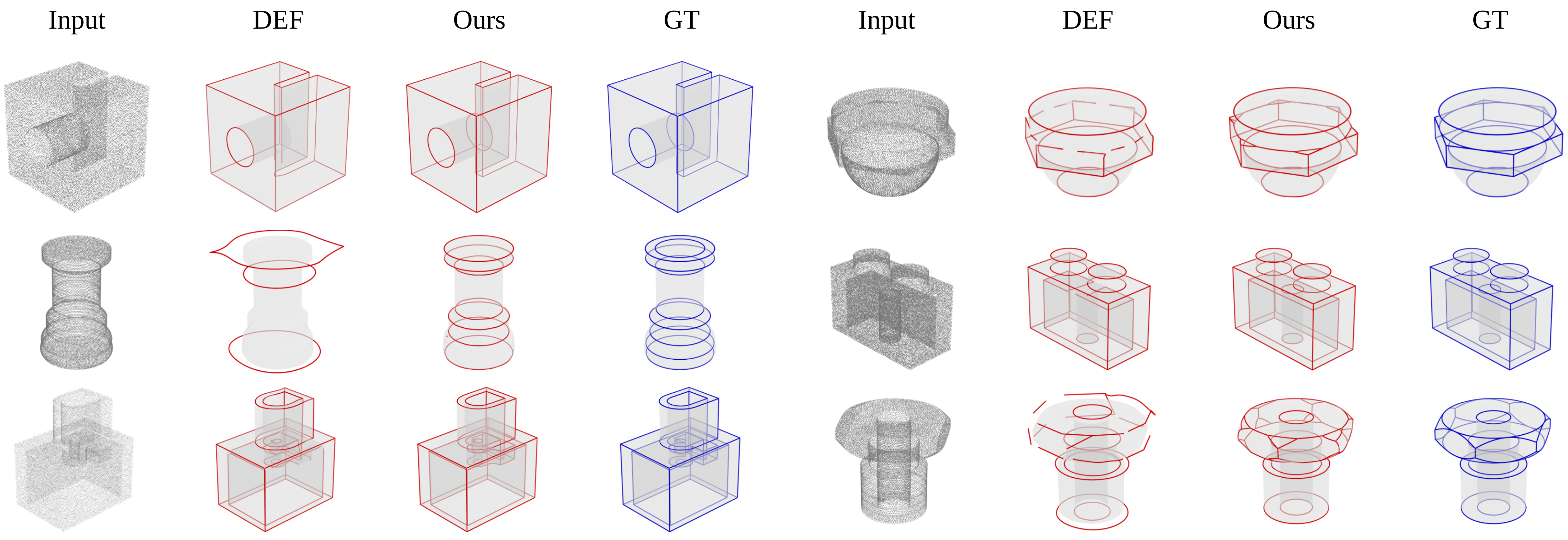}
	\caption{Qualitative comparisons with DEF on parametric curve extraction.}
	\label{fig:comp_paramcurve}
	\vspace{-5mm}
\end{figure*}

\subsection{Experiment Setup}
\noindent \textbf{Dataset.} We train and test our model on the ABC dataset~\cite{koch2019abc}, which has one million CAD models in total created by human users.  We use the first chunk for our experiments, which is already enough for use. Following~\cite{wang2020pie,matveev2022def,chen2022neural}, we consider \textit{sharp} curves in CAD models. However, some models need to be filtered due to data missing, shape repetition, or lack of sharp edges. The final filtered dataset contains 2364 models and is randomly split into a train set (80\%) and a test set (20\%). We provide additional details on dataset cleaning and data preparation in the supplemental.

\vspace{1mm}
\noindent \textbf{Implementation.} In our experiments, we set the resolution of NerVE grid at $32^3$. The effects of using higher resolutions are discussed in our ablation study (see Sec.~\ref{exp:ablation}). All input point clouds are normalized into $[-1,1]^3$. We train our networks on a NVIDIA RTX 3090 for 60 epochs. The Adam optimizer is used with an initial learning rate of 5e-4. We follow NDC~\cite{chen2022neural} and set the batch size as 1. The input point number varies for different shapes. On average, it is around 22,000. For more details like layer specifications and time cost, we provide them in the supplemental.

\vspace{1mm}
\noindent \textbf{Metrics.} We quantitatively evaluate the prediction of our networks using average recall \RecallCube{} and precision \PrecisionCube{} for edge occupancy, average correct rate \CorrectFace{} for edge orientations and average $L_2$ distance \DistancePoint{} for edge point positions. To define \CorrectFace{}, in the evaluation of edge orientation, a cube is regarded correct if all of its predictions of three faces are identical to GT, otherwise the cube is wrong. \CorrectFace{} is defined as the ratio of correct cubes in all considered cubes in an input. Note that the metrics are calculated under the cube masks defined in Sec.~\ref{sec:training}. 

To compare the quality of both PWL curves and parametric curves with other methods, we adopt the typical Chamfer distance (CD) and average Hausdorff distance (HD), 
which assess the similarity between two point sets. Assume $X,Y$ are two finite point sets, CD and HD can be calculated as:
\begin{align*}
    \text{CD} &= \frac{1}{|X|} \sum_{p \in X} \min_{q \in Y} \lVert p-q \rVert_2^2 + \frac{1}{|Y|} \sum_{q \in Y} \min_{p \in X} \lVert q-p \rVert_2^2,
\end{align*}
\begin{align*}
    \text{HD} &= \frac{1}{2} (\max_{p \in X} \min_{q \in Y} \lVert p-q \rVert_2 + \max_{q \in Y} \min_{p \in X} \lVert p-q \rVert_2).
\end{align*}

\subsection{Comparisons}
\noindent \textbf{Edge Estimation.} We evaluate our predicted PWL curves on our test set by comparing them with the baseline methods: VCM~\cite{merigot2010voronoi}, EC-NET~\cite{yu2018ec} and PIE-NET~\cite{wang2020pie}.
To measure the error between predicted edge points and the ground-truth, we convert PWL curves into point sets by sampling the midpoints of all edges. Numerical results are reported in Table~\ref{tab:comp_edgepoints} and visual comparisons are shown in Fig.~\ref{fig:comp_edgepoints_vis}. We provide the settings of baseline methods in the supplemental.

\begin{table}[htb]
    \centering
    \resizebox{0.86\columnwidth}{!}{
    \begin{tabular}{l|c|c|c|c}
    \toprule
           & VCM~\cite{merigot2010voronoi} & EC-NET~\cite{yu2018ec} & PIE-NET~\cite{wang2020pie} & Ours \\
        \hline
        CD$\downarrow$ & 0.0226 & 0.0037 & 0.0074 & \textbf{0.0012} \\
        \hline
        HD$\downarrow$ & 0.1941 & 0.1284 & 0.1318 & \textbf{0.0714} \\
    \bottomrule
    \end{tabular}}
    \caption{Quantitative comparisons on edge points estimation.}
    \label{tab:comp_edgepoints}
    \vspace{-4mm}
\end{table}
\vspace{1mm}

The quantitative and qualitative results show that our method significantly outperforms the baselines, where we produce accurate and uniformly distributed points visually. VCM and EC-NET are prone to generate redundant points around the ground-truth edges, thus producing higher CD and HD values. PIE-NET has a thinner band of outputs, but it tends to miss some edge points. 

\begin{figure*}[!t]
	\centering
	\includegraphics[width=1.\linewidth]{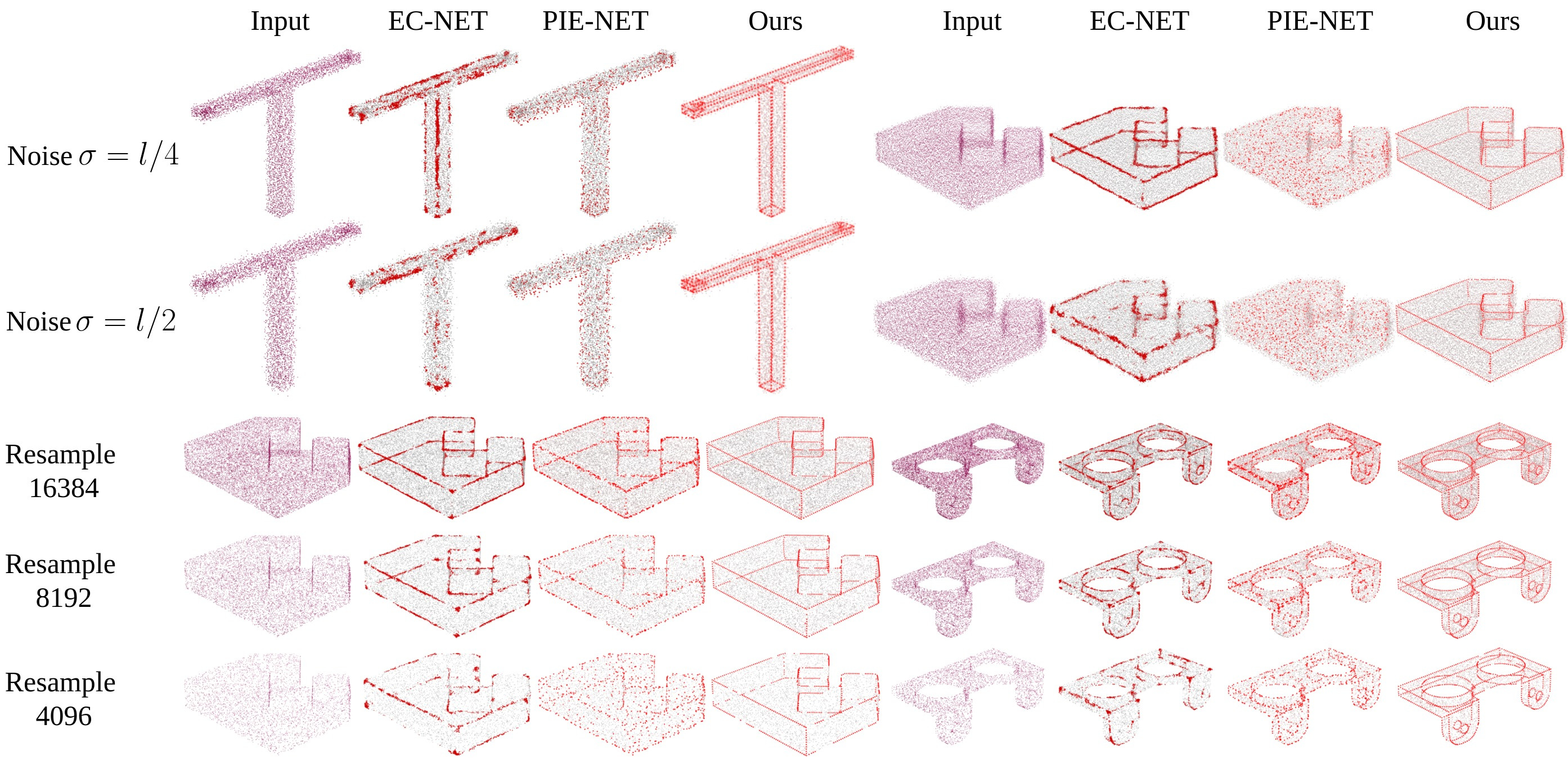}
	\caption{Qualitative comparisons with EC-NET~\cite{yu2018ec} and PIE-NET\cite{wang2020pie} on noisy input points or input points with varying sampling density. $l=2/32$ is the edge length of a cube in the grid.
	}
	\label{fig:stress_test}
	\vspace{-5mm}
\end{figure*}


\vspace{1mm}
\noindent \textbf{Parametric Curves.} With the predicted PWL curves as input, our method fit them to generate parametric curves.
We compare our parametric curves with the state-of-the-art method DEF~\cite{matveev2022def} both quantitatively and qualitatively on the DEF-Sim~\cite{matveev2022def} dataset, which has 68 carefully selected shapes from ABC dataset. Since the official code of DEF is not available, we use the input data in DEF-Sim and their results of parametric curves provided by DEF's authors. We directly use the reported results of PIE-Net and DEF from DEF's paper~\cite{matveev2022def} for quantitative comparisons (shown in Table~\ref{tab:comp_paramcurve}). Since PIE-NET does not provide the official implementation for its parametric curve inference, we only compare with DEF qualitatively in Fig.~\ref{fig:comp_paramcurve}.

\begin{table}[htb]
    \centering
    \resizebox{0.845\columnwidth}{!}{
    \begin{tabular}{l|c|c|c|c}
    \toprule
          & PIE-NET~\cite{wang2020pie} & DEF~\cite{matveev2022def} & Ours($32^3$) & Ours($64^3$) \\
        \hline
        CD$\downarrow$  & 0.97 & 0.04  & 0.008 & \textbf{0.005} \\
        \hline
        HD$\downarrow$   & 2.19 & 0.55  & 0.224 & \textbf{0.184}\\
    \bottomrule
    \end{tabular}}
    \caption{Quantitative comparisons on parametric curve extraction.}
    \label{tab:comp_paramcurve}
    \vspace{-4mm}
\end{table}
\vspace{1mm}


The results in Table~\ref{tab:comp_paramcurve} show that our method presents significantly better performance than DEF and PIE-NET, where the CD and HD errors of PIE-NET primarily arise from missing curve instances, which usually happen when keypoints (e.g., corner points or edge points) are wrongly predicted or simply missed. As shown in Fig.~\ref{fig:comp_paramcurve}, DEF performs well around sharp corners but struggles in processing structures with circle curves. In contrast, our method can handle closed curves and sharp corners consistently and produce convincing parametric curves in visual. 

\subsection{Stress Tests}
We investigate the robustness of our method using point clouds with varying noise or sampled point numbers. We train and test networks on our dataset with augmentation by adding Gaussian noise and random resampling. 

\vspace{1mm}
\noindent \textbf{NerVE Prediction.} Table~\ref{tab:stresstest_network} shows the effects of noise intensity and sampled point number on the prediction of our NerVE representation. We observe that even with a large noise intensity ($\sigma=l/2$, where $l$ is the edge length of a cube in the grid) or much fewer sampled points (4,096), our method still can produce reasonable results of PWL curves with low-level CD and HD errors.
\begin{table}[htb]
    \centering
    \resizebox{\columnwidth}{!}{
        \begin{tabular}{c|c|c|c|c|c|c|c}
        \toprule
            & & \RecallCube{}$\uparrow$ & \PrecisionCube{}$\uparrow$ & \CorrectFace{}$\uparrow$ & \DistancePoint{}$\downarrow$ & CD$\downarrow$ & HD$\downarrow$ \\
            \hline
            Clean & & \textbf{0.965} & \textbf{0.965} & \textbf{0.940} & \textbf{0.0030} & \textbf{0.0012} & \textbf{0.071} \\ 
            \hline
            \multirow{2}{2em}{Noise} & $\sigma=l/4$ & 0.919 & 0.937 & 0.887 & 0.0053 & 0.0019 & 0.097 \\
            & $\sigma=l/2$ & 0.892 & 0.923 & 0.854 & 0.0079 & 0.0027 & 0.110 \\
            \hline
            \multirow{3}{3em}{\#Sample Points} & 16384 & 0.952 & 0.955 & 0.919 & 0.0044 & 0.0019 & 0.095 \\
            & 8192 & 0.938 & 0.951 & 0.902 & 0.0048 & 0.0028 & 0.112 \\
            & 4096 & 0.923 & 0.941 & 0.876 & 0.0053 & 0.0057 & 0.146 \\
        \bottomrule
        \end{tabular}
    }
    \caption{Influence of input points with varying noise or sampled point numbers. $l=2/32$ is the edge length of a cube in the grid.}
    \label{tab:stresstest_network}
    \vspace{-4mm}
\end{table}
\vspace{1mm}

\noindent \textbf{Edge Estimation.} We also compare with VCM~\cite{merigot2010voronoi}, EC-NET~\cite{yu2018ec} and PIE-NET\cite{wang2020pie} on noisy or resampled inputs. Table~\ref{tab:stresstest_edgepoints} shows that our method achieves the best numerical performance. Fig.~\ref{fig:stress_test} further demonstrates that our method outperforms baseline methods and presents robustness against noisy or resampled inputs.

\begin{table}[htb]
    \centering
    \resizebox{\columnwidth}{!}{
        \begin{tabular}{c|c|c|c|c|c}
        \toprule
            & & VCM~\cite{merigot2010voronoi} & EC-NET~\cite{yu2018ec} & PIE-NET\cite{wang2020pie} & Ours \\
            \hline
            Clean & & 0.0226 & 0.0037 & 0.0074 & \textbf{0.0012} \\ 
            \hline
            \multirow{2}{2em}{Noise} & $\sigma=l/4$ & 0.0253 & 0.0055 & 0.0323 & \textbf{0.0019} \\
            & $\sigma=l/2$ & 0.0250 & 0.0132 & 0.0420 & \textbf{0.0027} \\
            \hline
            \multirow{3}{3em}{\#Sample Points} & 16384 & 0.0211 & 0.0039 & 0.0060 & \textbf{0.0019} \\
            & 8192 & 0.0214 & 0.0060 & 0.0145 & \textbf{0.0028} \\
            & 4096 & 0.0223 & 0.0130 & 0.0216 & \textbf{0.0057} \\
        \bottomrule
        \end{tabular}
    }
    \caption{Edge estimation errors (CD, the smaller the better) of four methods on noisy or resampled inputs. $l=2/32$ is the edge length of a cube in grid. Results of HD are provided in the supplemental. }
    \label{tab:stresstest_edgepoints}
    \vspace{-4mm}
\end{table}

\subsection{Ablation Study}
\label{exp:ablation}
\begin{figure}[ht]
	\centering
	\includegraphics[width=1.\linewidth]{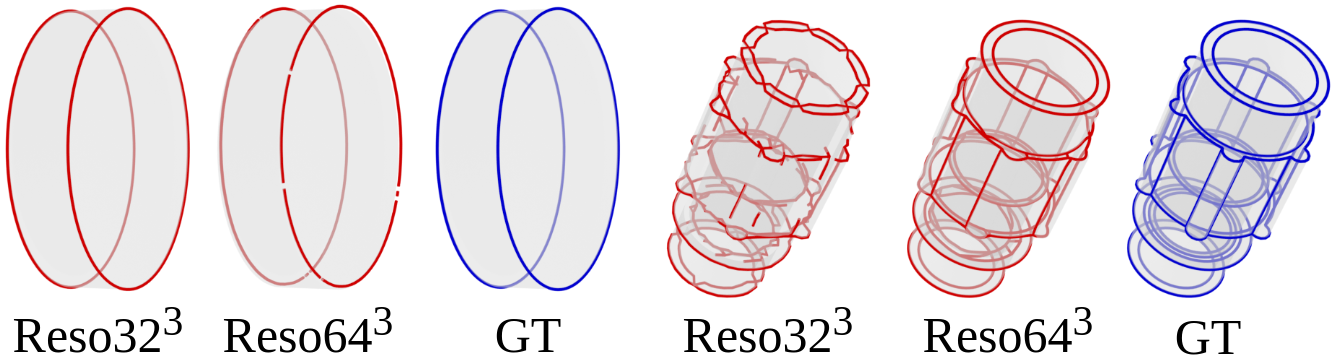}
	\caption{Qualitative comparisons on the resolution of $32^3$ and $64^3$ for \modelName{}. The resolution $64^3$ can represent complicated curves but also have imperfections in simple instances. }
	\label{fig:ablation_reso}
	\vspace{-3mm}
\end{figure}

\noindent \textbf{Resolution.} Table~\ref{tab:ablation_reso} shows the performance of our method under different NerVE grid resolution, where we observe that using resolution $32^3$ achieves slightly better performance on occupancy prediction of edge cubes. As a binary classification problem, data imbalance of edge occupancy is aggravated as resolution increases since the number of non-edge points grows faster than edge points number. Therefore, using resolution $64^3$ meets a more challenging classification problem, producing slightly worse \RecallCube{} and \PrecisionCube{} than using resolution $32^3$. Nevertheless, we notice that using resolution $64^3$ can achieve better CD and HD. Meanwhile, the parametric curves under resolution $64^3$ present better performance as shown in Table~\ref{tab:comp_paramcurve}. Fig.~\ref{fig:ablation_reso} shows the visualizations and provides some insights to explain the phenomenon.

\begin{table}[t]
    \centering
    \resizebox{0.875\columnwidth}{!}{
        \begin{tabular}{l|c|c|c|c|c|c}
        \toprule & \RecallCube{}$\uparrow$ & \PrecisionCube{}$\uparrow$ & \CorrectFace{}$\uparrow$ & \DistancePoint{}$\downarrow$ & CD$\downarrow$ & HD$\downarrow$ \\
            \hline
            Reso $32^3$ & \textbf{0.965}  & \textbf{0.965}  & \textbf{0.940}  & 0.003  & 0.0012  & 0.0714 \\
            \hline
            Reso $64^3$ & 0.958  & 0.960  & \textbf{0.940}  & \textbf{0.001}  & \textbf{0.0009}  & \textbf{0.0523} \\
        \bottomrule
        \end{tabular}
    }
    \caption{Influence of grid resolution on NerVE prediction and PWL curves generation. }
    \label{tab:ablation_reso}
    \vspace{-5mm}
\end{table}

\vspace{1mm}



For simple shapes, e.g., the left side of Fig.~\ref{fig:ablation_reso}, using a higher resolution (e.g., $64^3$) could disconnect at several positions on curves, which will degrade the edge occupancy accuracy of its PWL curves, but these artifacts (e.g., disconnection) can be well addressed by a simple post-processing in the following parametric curve extraction. On the contrary, using resolution $32^3$ suffers from representing complicated shapes (right side in Fig.~\ref{fig:ablation_reso}) while using resolution $64^3$ performs better and brings more geometric details. More experiments and analysis are provided in the supplemental. 


\noindent \textbf{Cube Point Choice.} As stated in Sec.~\ref{sec:method}, edge point position in a cube is defined as the midpoint of the truncated curve. Another option is to use a point which minimizes a quadratic error function (QEF), similar to the point definition in Dual Contouring~\cite{ju2002dual}. We provide the details in our supplemental. We validate our choice by restoring curves from ground-truth NerVE with two different definitions of point position. As shown in Table~\ref{tab:ablation_cubepoint_choice}, current definition of the point position clearly has better performance on curve restoration. 

\begin{table}[htb]
    \centering
    \resizebox{0.5\columnwidth}{!}{
        \begin{tabular}{l|c|c}
        \toprule
             & DC QEF & Our Choice \\
            \hline
            CD $\downarrow$ & 0.002799 & \textbf{0.000136} \\
            \hline
            HD $\downarrow$ & 0.052306 & \textbf{0.024263} \\
        \bottomrule
        \end{tabular}
    }
    \caption{Ablation study of edge point estimation on ground-truth NerVE. We measure the errors of generated PWL curves using two types of point positions in cubes, i.e, DC QEF~\cite{ju2002dual} and ours. }
    \label{tab:ablation_cubepoint_choice}
    \vspace{-4mm}
\end{table}

\vspace{1mm}
\begin{figure}[!t]
	\centering
	\includegraphics[width=0.75\linewidth]{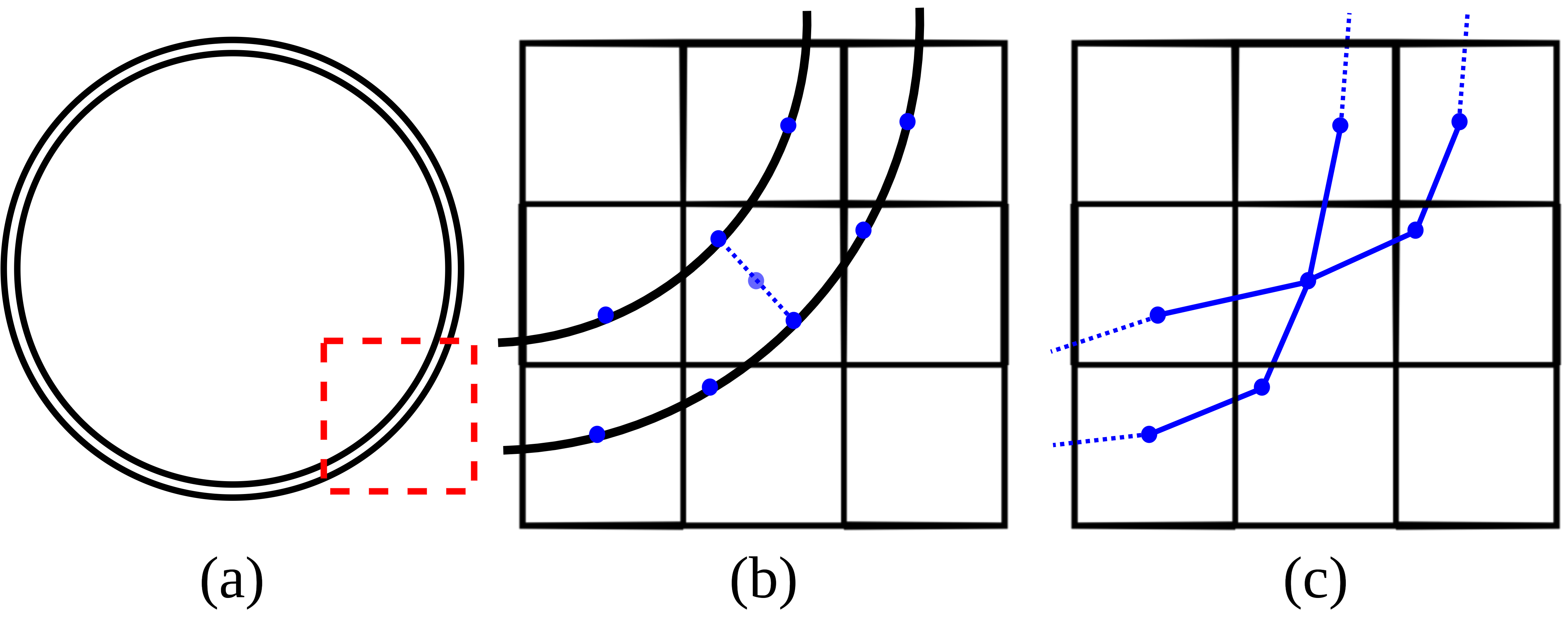}
	\caption{Limitations of our method. When two curves are close to each other (a), there is a possibility that two edge points respectively from them fall into the same cube, which would be processed into one edge point (c) with a low-resolution NerVE.}
	\label{fig:limitation}
	\vspace{-5mm}
\end{figure}

\section{Conclusions}
\label{sec:conclusion}
In this paper, we propose NerVE, a neural volumetric edge representation, for the extraction of parametric curves from a point cloud. This edge structure representation is fully compatible with the volumetric learning framework and can be easily converted to explicit PWL curves, which greatly reduces the complexity of following parametric curve fitting. The quantitative and qualitative results in the experiments evidently show the superiority of our method. 

\noindent \textbf{Limitations and Future Works.} One limitation is that our method may produce unexpected junction points. As illustrated in Figure~\ref{fig:limitation}, when two curves are so close and pass through the same cube, it will produce a junction point since we only predict one point in one cube by definition. This issue is essentially the cost of discretization, but it is insignificant in a statistical sense. In fact, there are only 1.62\% cubes with described junction points in all edge-occupied cubes in our dataset at resolution $32^3$, and the number decreases to 0.71\% as the resolution increases from $32^3$ to $64^3$. Another limitation is the vertex in one cube can only have at most 6 connections, due to the definition of \modelName{} and the structure of cube grids. In the future, we would like to devise a better representation that can completely solve these problems.

Designing a differentiable architecture for the extraction of PWL curves or parametric curves is also an interesting future work. Furthermore, notice \modelName{} can essentially represent general edges, we plan to test its ability boundary. For example, we wish to apply it not only to CAD models but also to some other shapes that have thin structures, such as the neural nerves of humans. 


\vspace{1mm}
\noindent\textbf{Acknowledgment.} 
This work was partially supported by NSFC-62172348, the Basic Research Project No. HZQB-KCZYZ-2021067 of Hetao Shenzhen-HK S\&T Cooperation Zone, the National Key R\&D Program of China with grant No. 2018YFB1800800, by Shenzhen Outstanding Talents Training Fund 202002, by Guangdong Research Projects No. 2017ZT07X152 and No. 2019CX01X104, by the Guangdong Provincial Key Laboratory of Future Networks of Intelligence (Grant No. 2022B1212010001), and by Shenzhen Key Laboratory of Big Data and Artificial Intelligence (Grant No. ZDSYS201707251409055). It was also supported in part by Outstanding Yound Fund of Guangdong Province with No.  2023B1515020055 and Shenzhen General Project with No. JCYJ20220530143604010. It was also sponsored by CCF-Tencent Open Research Fund.

{\small
\bibliographystyle{ieee_fullname}
\bibliography{egbib}

\begin{thebibliography}{10}\itemsep=-1pt

\bibitem{bazazian2015fast}
Dena Bazazian, Josep~R Casas, and Javier Ruiz-Hidalgo.
\newblock Fast and robust edge extraction in unorganized point clouds.
\newblock In {\em 2015 international conference on digital image computing:
  techniques and applications (DICTA)}, pages 1--8. IEEE, 2015.

\bibitem{bazazian2021edc}
Dena Bazazian and M~Eul{\`a}lia Par{\'e}s.
\newblock Edc-net: Edge detection capsule network for 3d point clouds.
\newblock {\em Applied Sciences}, 11(4):1833, 2021.

\bibitem{chen2022neural}
Zhiqin Chen, Andrea Tagliasacchi, Thomas Funkhouser, and Hao Zhang.
\newblock Neural dual contouring.
\newblock {\em arXiv preprint arXiv:2202.01999}, 2022.

\bibitem{chen2019learning}
Zhiqin Chen and Hao Zhang.
\newblock Learning implicit fields for generative shape modeling.
\newblock In {\em Proceedings of the IEEE Conference on Computer Vision and
  Pattern Recognition}, pages 5939--5948, 2019.

\bibitem{chen2021neural}
Zhiqin Chen and Hao Zhang.
\newblock Neural marching cubes.
\newblock {\em ACM Transactions on Graphics (TOG)}, 40(6):1--15, 2021.

\bibitem{choy20163d}
Christopher~B Choy, Danfei Xu, JunYoung Gwak, Kevin Chen, and Silvio Savarese.
\newblock 3d-r2n2: A unified approach for single and multi-view 3d object
  reconstruction.
\newblock In {\em European conference on computer vision}, pages 628--644.
  Springer, 2016.

\bibitem{daniels2007robust}
Joel~II Daniels, Linh~K Ha, Tilo Ochotta, and Claudio~T Silva.
\newblock Robust smooth feature extraction from point clouds.
\newblock In {\em IEEE International Conference on Shape Modeling and
  Applications 2007 (SMI'07)}, pages 123--136. IEEE, 2007.

\bibitem{demarsin2006detection}
Kris Demarsin, Denis Vanderstraeten, Tim Volodine, and Dirk Roose.
\newblock Detection of closed sharp feature lines in point clouds for reverse
  engineering applications.
\newblock In {\em International Conference on Geometric Modeling and
  Processing}, pages 571--577. Springer, 2006.

\bibitem{demarsin2007detection}
Kris Demarsin, Denis Vanderstraeten, Tim Volodine, and Dirk Roose.
\newblock Detection of closed sharp edges in point clouds using normal
  estimation and graph theory.
\newblock {\em Computer-Aided Design}, 39(4):276--283, 2007.

\bibitem{fabri2009cgal}
Andreas Fabri and Sylvain Pion.
\newblock Cgal: The computational geometry algorithms library.
\newblock In {\em Proceedings of the 17th ACM SIGSPATIAL international
  conference on advances in geographic information systems}, pages 538--539,
  2009.

\bibitem{fan2017point}
Haoqiang Fan, Hao Su, and Leonidas~J Guibas.
\newblock A point set generation network for 3d object reconstruction from a
  single image.
\newblock In {\em Proceedings of the IEEE conference on computer vision and
  pattern recognition}, pages 605--613, 2017.

\bibitem{fleishman2005robust}
Shachar Fleishman, Daniel Cohen-Or, and Cl{\'a}udio~T Silva.
\newblock Robust moving least-squares fitting with sharp features.
\newblock {\em ACM transactions on graphics (TOG)}, 24(3):544--552, 2005.

\bibitem{groueix2018papier}
Thibault Groueix, Matthew Fisher, Vladimir~G Kim, Bryan~C Russell, and Mathieu
  Aubry.
\newblock A papier-m{\^a}ch{\'e} approach to learning 3d surface generation.
\newblock In {\em Proceedings of the IEEE conference on computer vision and
  pattern recognition}, pages 216--224, 2018.

\bibitem{gumhold2001feature}
Stefan Gumhold, Xinlong Wang, Rob~S MacLeod, et~al.
\newblock Feature extraction from point clouds.
\newblock In {\em IMR}, pages 293--305, 2001.

\bibitem{guo2022complexgen}
Haoxiang Guo, Shilin Liu, Hao Pan, Yang Liu, Xin Tong, and Baining Guo.
\newblock Complexgen: Cad reconstruction by b-rep chain complex generation.
\newblock {\em ACM Transactions on Graphics (TOG)}, 41(4):1--18, 2022.

\bibitem{hackel2016contour}
Timo Hackel, Jan~D Wegner, and Konrad Schindler.
\newblock Contour detection in unstructured 3d point clouds.
\newblock In {\em Proceedings of the IEEE conference on computer vision and
  pattern recognition}, pages 1610--1618, 2016.

\bibitem{himeur2021pcednet}
Chems-Eddine Himeur, Thibault Lejemble, Thomas Pellegrini, Mathias Paulin, Loic
  Barthe, and Nicolas Mellado.
\newblock Pcednet: A lightweight neural network for fast and interactive edge
  detection in 3d point clouds.
\newblock {\em ACM Transactions on Graphics (TOG)}, 41(1):1--21, 2021.

\bibitem{hofmann1989geometric}
Christoph~M Hofmann.
\newblock {\em Geometric and solid modeling: an introduction}.
\newblock Morgan Kaufmann, 1989.

\bibitem{ju2002dual}
Tao Ju, Frank Losasso, Scott Schaefer, and Joe Warren.
\newblock Dual contouring of hermite data.
\newblock In {\em Proceedings of the 29th annual conference on Computer
  graphics and interactive techniques}, pages 339--346, 2002.

\bibitem{koch2019abc}
Sebastian Koch, Albert Matveev, Zhongshi Jiang, Francis Williams, Alexey
  Artemov, Evgeny Burnaev, Marc Alexa, Denis Zorin, and Daniele Panozzo.
\newblock Abc: A big cad model dataset for geometric deep learning.
\newblock In {\em Proceedings of the IEEE/CVF Conference on Computer Vision and
  Pattern Recognition}, pages 9601--9611, 2019.

\bibitem{liao2018deep}
Yiyi Liao, Simon Donne, and Andreas Geiger.
\newblock Deep marching cubes: Learning explicit surface representations.
\newblock In {\em Proceedings of the IEEE Conference on Computer Vision and
  Pattern Recognition}, pages 2916--2925, 2018.

\bibitem{lin2015line}
Yangbin Lin, Cheng Wang, Jun Cheng, Bili Chen, Fukai Jia, Zhonggui Chen, and
  Jonathan Li.
\newblock Line segment extraction for large scale unorganized point clouds.
\newblock {\em ISPRS Journal of Photogrammetry and Remote Sensing},
  102:172--183, 2015.

\bibitem{liu2021pc2wf}
Yujia Liu, Stefano D'Aronco, Konrad Schindler, and Jan~D Wegner.
\newblock Pc2wf: 3d wireframe reconstruction from raw point clouds.
\newblock In {\em International Conference on Learning Representations (ICLR)},
  2021.

\bibitem{lorensen1987marching}
William~E Lorensen and Harvey~E Cline.
\newblock Marching cubes: A high resolution 3d surface construction algorithm.
\newblock {\em ACM siggraph computer graphics}, 21(4):163--169, 1987.

\bibitem{matveev20213d}
Albert Matveev, Alexey Artemov, Denis Zorin, and Evgeny Burnaev.
\newblock 3d parametric wireframe extraction based on distance fields.
\newblock In {\em 2021 4th International Conference on Artificial Intelligence
  and Pattern Recognition}, pages 316--322, 2021.

\bibitem{matveev2022def}
Albert Matveev, Ruslan Rakhimov, Alexey Artemov, Gleb Bobrovskikh, Vage
  Egiazarian, Emil Bogomolov, Daniele Panozzo, Denis Zorin, and Evgeny Burnaev.
\newblock Def: Deep estimation of sharp geometric features in 3d shapes.
\newblock {\em ACM Transactions on Graphics (TOG)}, 41(4):1--22, 2022.

\bibitem{merigot2010voronoi}
Quentin M{\'e}rigot, Maks Ovsjanikov, and Leonidas~J Guibas.
\newblock Voronoi-based curvature and feature estimation from point clouds.
\newblock {\em IEEE Transactions on Visualization and Computer Graphics},
  17(6):743--756, 2010.

\bibitem{mescheder2019occupancy}
Lars Mescheder, Michael Oechsle, Michael Niemeyer, Sebastian Nowozin, and
  Andreas Geiger.
\newblock Occupancy networks: Learning 3d reconstruction in function space.
\newblock In {\em Proceedings of the IEEE Conference on Computer Vision and
  Pattern Recognition}, pages 4460--4470, 2019.

\bibitem{ni2016edge}
Huan Ni, Xiangguo Lin, Xiaogang Ning, and Jixian Zhang.
\newblock Edge detection and feature line tracing in 3d-point clouds by
  analyzing geometric properties of neighborhoods.
\newblock {\em Remote Sensing}, 8(9):710, 2016.

\bibitem{nie2016extracting}
Jianhui Nie.
\newblock Extracting feature lines from point clouds based on smooth shrink and
  iterative thinning.
\newblock {\em Graphical Models}, 84:38--49, 2016.

\bibitem{park2019deepsdf}
Jeong~Joon Park, Peter Florence, Julian Straub, Richard Newcombe, and Steven
  Lovegrove.
\newblock Deepsdf: Learning continuous signed distance functions for shape
  representation.
\newblock In {\em Proceedings of the IEEE Conference on Computer Vision and
  Pattern Recognition}, pages 165--174, 2019.

\bibitem{pauly2003multi}
Mark Pauly, Richard Keiser, and Markus Gross.
\newblock Multi-scale feature extraction on point-sampled surfaces.
\newblock In {\em Computer graphics forum}, pages 281--289. Wiley Online
  Library, 2003.

\bibitem{qi2017pointnet++}
Charles~Ruizhongtai Qi, Li Yi, Hao Su, and Leonidas~J Guibas.
\newblock Pointnet++: Deep hierarchical feature learning on point sets in a
  metric space.
\newblock {\em Advances in neural information processing systems}, 30, 2017.

\bibitem{stroud2006boundary}
Ian Stroud.
\newblock {\em Boundary representation modelling techniques}.
\newblock Springer Science \& Business Media, 2006.

\bibitem{tan2022coarse}
Xuefeng Tan, Dejun Zhang, Long Tian, Yiqi Wu, and Yilin Chen.
\newblock Coarse-to-fine pipeline for 3d wireframe reconstruction from point
  cloud.
\newblock {\em Computers \& Graphics}, 106:288--298, 2022.

\bibitem{wang2018pixel2mesh}
Nanyang Wang, Yinda Zhang, Zhuwen Li, Yanwei Fu, Wei Liu, and Yu-Gang Jiang.
\newblock Pixel2mesh: Generating 3d mesh models from single rgb images.
\newblock In {\em Proceedings of the European Conference on Computer Vision
  (ECCV)}, pages 52--67, 2018.

\bibitem{wang2020pie}
Xiaogang Wang, Yuelang Xu, Kai Xu, Andrea Tagliasacchi, Bin Zhou, Ali
  Mahdavi-Amiri, and Hao Zhang.
\newblock Pie-net: Parametric inference of point cloud edges.
\newblock {\em Advances in neural information processing systems},
  33:20167--20178, 2020.

\bibitem{weber2010sharp}
Christopher Weber, Stefanie Hahmann, and Hans Hagen.
\newblock Sharp feature detection in point clouds.
\newblock In {\em 2010 Shape Modeling International Conference}, pages
  175--186. IEEE, 2010.

\bibitem{weber2011methods}
Christopher Weber, Stefanie Hahmann, and Hans Hagen.
\newblock Methods for feature detection in point clouds.
\newblock In {\em Visualization of Large and Unstructured Data
  Sets-Applications in Geospatial Planning, Modeling and Engineering (IRTG 1131
  Workshop)}. Schloss Dagstuhl-Leibniz-Zentrum fuer Informatik, 2011.

\bibitem{weiler1986topological}
Kevin~J Weiler.
\newblock {\em Topological structures for geometric modeling (Boundary
  representation, manifold, radial edge structure)}.
\newblock Rensselaer Polytechnic Institute, 1986.

\bibitem{xia2017fast}
Shaobo Xia and Ruisheng Wang.
\newblock A fast edge extraction method for mobile lidar point clouds.
\newblock {\em IEEE Geoscience and Remote Sensing Letters}, 14(8):1288--1292,
  2017.

\bibitem{yu2018ec}
Lequan Yu, Xianzhi Li, Chi-Wing Fu, Daniel Cohen-Or, and Pheng-Ann Heng.
\newblock Ec-net: an edge-aware point set consolidation network.
\newblock In {\em Proceedings of the European conference on computer vision
  (ECCV)}, pages 386--402, 2018.

\end{thebibliography}
}

\clearpage

\appendix

In this supplementary material, we provide additional details of our method in Sec.~\ref{sec:supp_method}, data collection setup in Sec.~\ref{sec:supp_dataset}, and additional ablation study, numerical and visual results in Sec.~\ref{sec:supp_exp}.

\section{Details of Method}
\label{sec:supp_method}
In this section, we present network specifications for learning \modelName{} and list the post-processing details for the PWL curves refinement and the final parametric curve fitting.

\subsection{Network Details}

\paragraph{Encoder.} We use a simplified dense PointNet++~\cite{qi2017pointnet++} as our encoder. Specifically, given a point cloud of shape $(N,3)$ ($N$ is the point number), we first find k-nearest neighbors ($k=8$ as in~\cite{chen2022neural}) of each point and reform them into a tensor of shape $(N,8,3)$ as input. Then we apply a network consisting of 4 layers of MLP, where the latent size is $128$ and the output shape is $(N,8,128)$. We finally obtain point features of $(N,128)$ by using a max pooling function in the second dimension. After that, the point features are fused as cube features $(32,32,32,128)$ (32 is the grid resolution) by average pooling when multiple points appear in the same cube. Three 3D-convolution layers are then applied to the cube features, where the kernel size, stride, and padding are 3, 1, and 1 respectively. 
We use Leaky ReLU as the activation function. The final shape of the feature grid is $(32,32,32,128)$.

\paragraph{Decoder.}We adopt a 5-layer MLP as the decoder to predict edge occupancy, orientations, and edge point position. The latent layer size of the MLP is $128$. Each cube in the feature grid has a feature size of $128$, which is directly decoded by the occupancy decoder into a one-dimensional scalar. It is decoded by a position decoder into three floats. For the orientations decoder, as shown in Figure 2 in our paper, we first concatenate the cube feature with its three adjacent cube features, which means the input to the orientations decoder is of shape $(3, 128+128)$. The orientations decoder takes the concatenated feature as the input and produces three one-dimensional scalars.
Sigmoid activation is used as the last layer in both the occupancy decoder and the orientations decoder. 
The output of our position decoder is clipped into $[-1,1]^3$ to be consistent with the coordinate system of the input point cloud (see \emph{Input Point Cloud Pre-processing} in Sec.~\ref{sec:supp_dataset} for more details on coordinate transform).

\subsection{Parametric Curve Extraction}
We introduce the processing details in parametric curve extraction, including the refinement of PWL curves and parametric fitting.
Note that the whole procedure of extraction is fast, where the parametric curve extraction takes only 0.018 seconds per shape on average from all shapes in the test set (472 shapes), where the average vertex number on shape curves is about 22000.

\paragraph{PWL Curves Refinement.}To refine the predicted PWL curves from our network for parametric curve extraction, we propose several post-processing steps. 
In the following settings, the predicted PWL curves are regarded as an undirected graph, where the definition of vertex degree is the same as in general graph theory. 

\begin{figure}[ht]
	\centering
	\includegraphics[width=0.75\linewidth]{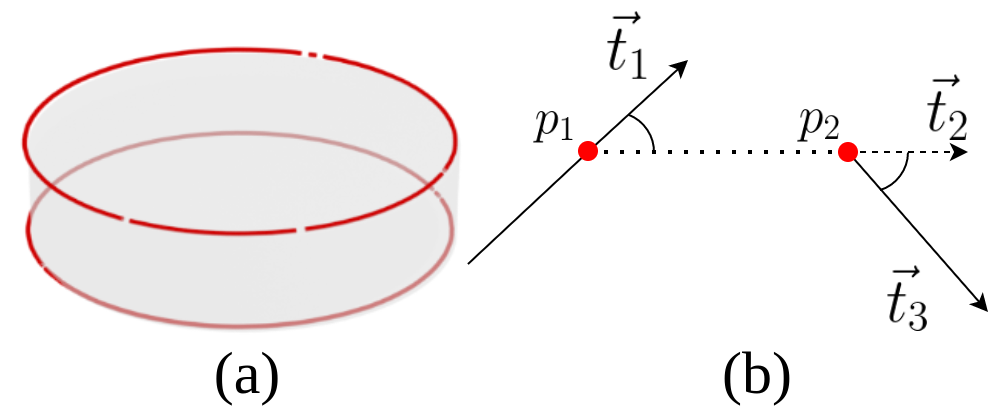}
	\caption{(a) An example that requires curve reconnection. (b) Let $p_1,p_2$ be two close \DegreeOne vertices. $\Vec{t}_1, \Vec{t}_3$ are tangent vectors for $p_1, p_2$ (defined as in the PWL curves) and $\Vec{t}_2 = \protect\overrightarrow{p_1p_2} / \lVert \protect\overrightarrow{p_1p_2}\rVert$.
	}
	\label{fig:post_reconnection}
	\vspace{-2mm}
\end{figure}
\noindent \textbf{Step 1: Point Reconnection.}
We first find all vertices with degree 1 and denote such vertices as \DegreeOne vertices. Then we add an edge between two \DegreeOne vertices if the distance of these vertices is smaller than a given threshold $\delta_r$ meanwhile, their tangent vectors should be consistent enough. See Fig.~\ref{fig:post_reconnection} for an illustration of reconnection as well as the definition of consistency between tangent vectors. Take Fig.~\ref{fig:post_reconnection} (b) as an example, the consistency of tangent vectors is defined by $\Vec{t}_1 \cdot \Vec{t}_2 + \Vec{t}_2 \cdot \Vec{t}_3$. If $\lVert \overrightarrow{p_1p_2}\rVert < \delta_r$ and $\Vec{t}_1 \cdot \Vec{t}_2 + \Vec{t}_2 \cdot \Vec{t}_3 > \sqrt{2}$ ($\sqrt{2}$ is fixed, which works well in all our experiments), we can connect $p_1$ and $p_2$. 

\begin{figure}[htb]
	\centering
	\includegraphics[width=0.875\linewidth]{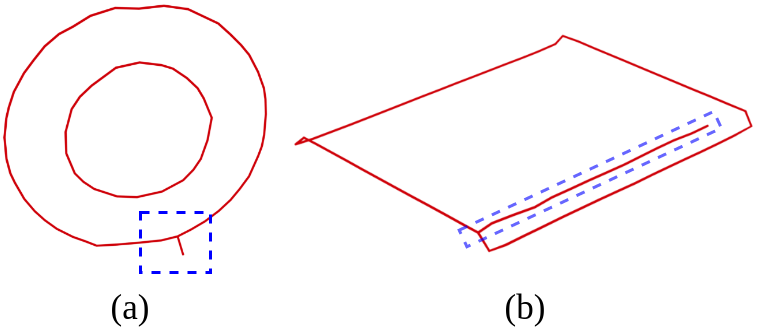}
	\caption{Examples of extra edges. (a) A short protruding edge will be removed if its vertex number is less than $N_p$ ($N_p=5$ in our experiments). (b) A long extra edge is also removed because of the B-Rep constraints. Such long extra edges usually appear around non-sharp edges of shapes (referring to Fig.~\ref{fig:abc_sharp_edges}), and it is difficult for neural networks to distinguish them from sharp edges.
	}
	\label{fig:post_extra_line}
	\vspace{-2mm}
\end{figure}
\noindent \textbf{Step 2: Extra Edges Removal.}
As shown in Fig.~\ref{fig:post_extra_line}, there could be extra edge segments in the space. To remove these outliers, we first find all paths which start with a \DegreeOne vertex. A path on the PWL curves graph is defined by adding constraints: $\deg (V_{\text{in}}) =2, \deg (V_{\text{end}}) \neq 2$, where $V_{\text{in}}$ is interior vertex and $V_{\text{end}}$ is the end vertex of the path and $\deg(V)$ means the degree of vertex $V$ on PWL curves graph. 
Then we remove one from these paths if its vertex number is less than $N_p$. In this way, short extra edges can be removed, and there remain long extra edges that failed to reconnect in \emph{Step 1}. 
We provide an option on keeping long protruding edges according to users' preference on conforming to the B-Rep constraints.
In particular, long protruding edges are removed if B-Rep constraints are strictly required and kept otherwise.
In our experiments, we choose to impose the B-Rep constraints for final parametric curves and remove such long protruding edges.

\begin{figure}[ht]
	\centering
	\includegraphics[width=0.675\linewidth]{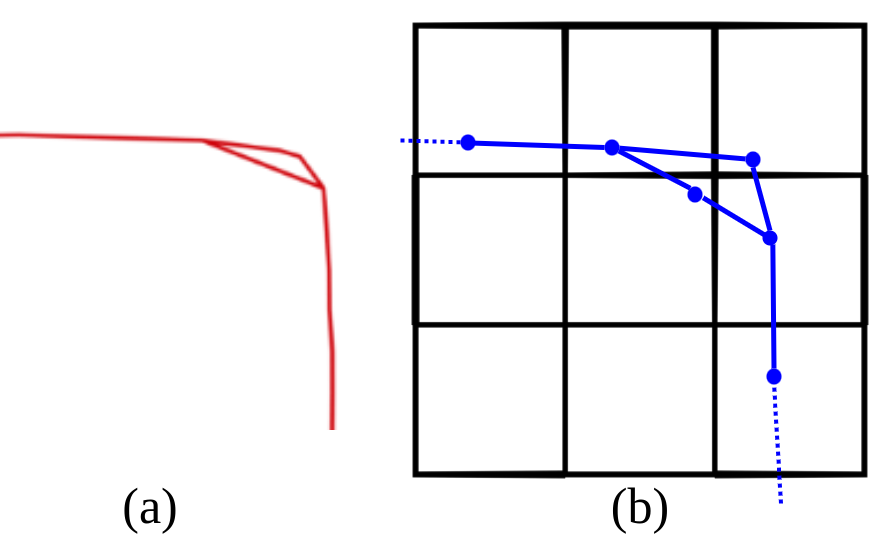}
	\caption{(a) There are multiple paths between two degree$>2$ vertices(endpoints). (b) 2D illustration for generation of multi-paths. When a curve is near the edges or faces of the cube, the network might predict the additional point, which can cause multiple paths between two endpoints.
	}
	\label{fig:post_multipath}
	\vspace{-4mm}
\end{figure}
\noindent \textbf{Step 3: Multi-paths Handling.}
In our method, all vertices with degree $>2$ are regarded as endpoints of curves. However, there could be multiple close paths between two endpoints as shown in Fig.~\ref{fig:post_multipath}, When curves are close to edges or faces of the cube, the network might predict the additional point causing multiple paths. To handle this issue, we can randomly select one of the paths and delete others since all paths are close in geometry, and we finally fit the paths in the sense of least squares. Two paths from the same pair of endpoints are defined to be close if the Chamfer distance of these two paths is lower than a given threshold $\delta_p$. We repeat this process until all extra paths are eliminated. 

\begin{table}[htb]
    \centering
    \resizebox{0.865\columnwidth}{!}{
        \begin{tabular}{c|c|c|c|c|c}
        \toprule
            \multirow{3}{0em}{}
            $\delta_r$ & Value & $2l$ & $3l$ & $4l$ & $5l$ \\
            \hline
            & CD$\downarrow$ & 0.0112  & 0.0066  & \textbf{0.0051}  & 0.0121 \\
            & HD$\downarrow$ & 0.2246  & 0.1924  & \textbf{0.1844}  & 0.2870 \\
            \hline
            \multirow{3}{0em}{} 
            $N_p$ & Value & $3$ & $4$ & $5$ & $6$ \\
            \hline
            & CD$\downarrow$ & 0.0054  & 0.0054  & \textbf{0.0051}  & 0.0061 \\
            & HD$\downarrow$ & 0.2046  & 0.2010  & \textbf{0.1844}  & 0.1847 \\
            \hline
            \multirow{3}{0em}{}
            $\delta_p$ & Value & $0.5l$ & $2l$ & $4l$ & $\infty$ \\
            \hline
            & CD$\downarrow$ & 0.00509  & \textbf{0.00508}  & 0.00509  & 0.00512 \\
            & HD$\downarrow$ & 0.18442  & 0.18442  & 0.18442  & 0.18442 \\
        \bottomrule
        \end{tabular}
    }
    \caption{Different CD and HD errors of final parametric curves when $\delta_r, N_p, \delta_p$ taking different values. $l=2/r$ is the edge length of a cube in the grid ($r=64$ for resolution $64^3$). By default, $\delta_r=4l, N_p=5, \delta_p=2l$. In the experiments, one parameter value changes and other parameters remain the default values. For $\delta_p$, the value of $\infty$ means we handle all possible multi-paths cases without checking the Chamfer distance.
    }
    \label{tab:Supp_parameters}
    \vspace{-4mm}
\end{table}
\paragraph{Choices of Parameters.}Three parameters are discussed in PWL curves refinement: $\delta_r, N_p,\delta_p$. To choose appropriate values, we compare the CD and HD errors of the final parametric curves in different parameter settings, as shown in Table~\ref{tab:Supp_parameters}. Based on the comparison, we choose $\delta_r=4l, N_p=5, \delta_p=2l$, where $l=2/r$ ($r=64$ for resolution $64^3$) is the edge length of a cube in the grid.

\paragraph{Parametric Curve Fitting.}After obtaining paths between pairs of endpoints or closed paths, we can use an off-the-shelf spline fitting library for parametric curve fitting on these paths. Specifically, we use the function \emph{make\_lsq\_spline} from \emph{Scipy}, which can fit given points in the sense of least squares with BSpline functions. For our setting of BSpline functions, the order of spline is $3$; the number of knots (except the knots for start and end points) is half of the number of path vertices; knots are uniformly sampled in $[0,1]$. 
Note that the positions of all endpoints are fixed during fitting.

For closed paths, we first try direct 3D circle fitting on one closed path since most closed curves in the ABC dataset are circles~\cite{wang2020pie}. If the fitting error is large, which means the closed path is probably not a circle, we simply apply the previous spline fitting to it. Here, the threshold for the fitting error is fixed as $0.001$, which works fine in all our experiments. For 3D circle fitting, we use a feasible and simple approach. We first use principal component analysis (PCA) on points and then convert it to a 2D circle fitting problem, which can be easily solved, finally we map the fitted 2D circle into $\mathbb{R}^3$ by PCA and obtain the 3D circle. 
\section{Details of Dataset}
\label{sec:supp_dataset}
In this section, we provide more details about data processing, including data cleaning of raw ABC dataset~\cite{koch2019abc}, ground truth data preparation of NerVE attributes, and pre-processing for input point clouds.
\begin{figure}[ht]
	\centering
	\includegraphics[width=0.9\linewidth]{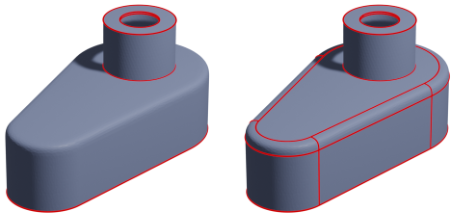}
	\caption{Differences between sharp edge and non-sharp edge. We highlight the sharp edges on the left figure, while highlight both sharp and non-sharp edges on the right figure.
	}
	\label{fig:abc_sharp_edges}
	\vspace{-5mm}
\end{figure}
\paragraph{Data Cleaning.}As discussed in our paper, the dataset has to be cleaned due to data missing, shape repetition, and lack of sharp edges. For those shapes with little difference in geometry or structure, we regard them as repeated. As for the lack of sharp edges, there are cases where a shape does not possess any sharp edges (e.g. a sphere) or only has a few sharp edges. Here we use the sharp edges originally defined in the ABC dataset~\cite{koch2019abc} and Fig.~\ref{fig:abc_sharp_edges} shows the difference. To filter out the mentioned shapes, we manually examined all shapes from the first chunk and obtained 2,364 shapes for network learning. 

\paragraph{Ground Truth Preparation.}To obtain the ground truth of cube attributes in NerVE, we first obtain the PWL curves from the GT parametric curves by uniform sampling and denote the PWL curves as the GT PWL curves. In a grid of $32^3$ cubes (here we assume the grid resolution is $32$), all cubes intersected with the GT PWL curves are labeled True, which are occupied cubes. 
Similarly, all faces intersected with the GT PWL curves are labeled True. Intersections of GT PWL curves with cubes can be easily calculated by considering each line segment in GT PWL curves. In the occupied cubes, we take the midpoint of the truncated GT PWL curve inside the cube as the point position.

\paragraph{Input Point Cloud Pre-processing.} To benefit network training, we normalize the input point cloud and GT point positions of cubes. For the input point cloud, we first subtract the point position from its $k$ nearest neighbors for each point, then multiply by a factor of $r$ ($r=32$ for resolution $32^3$). For the GT global point positions, we convert it to the local coordinates of its cube, where the center of a cube is the new origin, and the axes are scaled by $r$ ($r=32$ for resolution $32^3$). In this way, the range of the GT point position becomes $[-1,1]^3$.
\section{More Details of Experiments}
\label{sec:supp_exp}
In this section, we give more details about the experiments, such as baseline settings in the comparison and a detailed explanation of cube point choice in the ablation study. Finally, we show more results of our method in Fig.~\ref{fig:more_results}. 

\paragraph{Baseline Settings.} 
In our experiments, we adopt VCM~\cite{merigot2010voronoi}, EC-NET~\cite{yu2018ec}, and PIE-NET~\cite{wang2020pie} as baselines to evaluate our proposed method. Specifically, we use the implementation of VCM in the CGAL library~\cite{fabri2009cgal}. Given a point cloud for testing, we compute the Voronoi covariance at each point, where the offset radius is 0.2, the convolution radius is 0.25, and the pareto-optimal threshold is 0.24. When a Voronoi covariance value is larger than the threshold, we consider the corresponding point to belong to an edge. For EC-NET and PIE-NET, we utilize the released source codes and pre-trained models to test the input point cloud with their specified normalization, and then transform the outputs to align with the ground truth for a fair evaluation.

\paragraph{Time Consumption.}
The average inference times of VCM, EC-Net, PIE-Net, and Ours are 2.06, 0.84, 0.52, and 0.15 seconds, respectively. For post-processing, our method takes 0.02s on average, which is more efficient than the post-processing of PIE-Net (3.01s). It can be explained by using masks, which can only choose surface cubes for the calculation to reduce consumption and make it more efficient than other approaches. 

\begin{table}[htb]
    \centering
    \resizebox{\columnwidth}{!}{
        \begin{tabular}{c|c|c|c|c|c}
        \toprule
            & & VCM~\cite{merigot2010voronoi} & EC-NET~\cite{yu2018ec} & PIE-NET\cite{wang2020pie} & Ours \\
            \hline
            Clean & & 0.194 & 0.128 & 0.132 & \textbf{0.071} \\ 
            \hline
            \multirow{2}{2em}{Noise} & $\sigma=l/4$ & 0.200 & 0.222 & 0.289 & \textbf{0.097} \\
            & $\sigma=l/2$ & 0.270 & 0.278 & 0.301 & \textbf{0.110} \\
            \hline
            \multirow{3}{3em}{\#Sample Points} & 16384 & 0.185 & 0.135 & 0.179 & \textbf{0.095}  \\
            & 8192 & 0.192 & 0.164 & 0.238 & \textbf{0.112} \\
            & 4096 & 0.203 & 0.212 & 0.246 & \textbf{0.146} \\
        \bottomrule
        \end{tabular}
    }
    \caption{Edge estimation errors (HD, the smaller the better) of four methods on noisy or resampled inputs. $l=2/32$ is the edge length of a cube in grid.}
    \label{tab:Supp_stresstest_edgepoints}
    \vspace{-4mm}
\end{table}

\paragraph{HD Results of Stress Tests.} Table~\ref{tab:Supp_stresstest_edgepoints} shows additional quantitative results of edge estimation in stress tests. Here the metric is HD instead of CD used in the main paper.

\begin{figure}[tb]
	\centering
	\includegraphics[width=\linewidth]{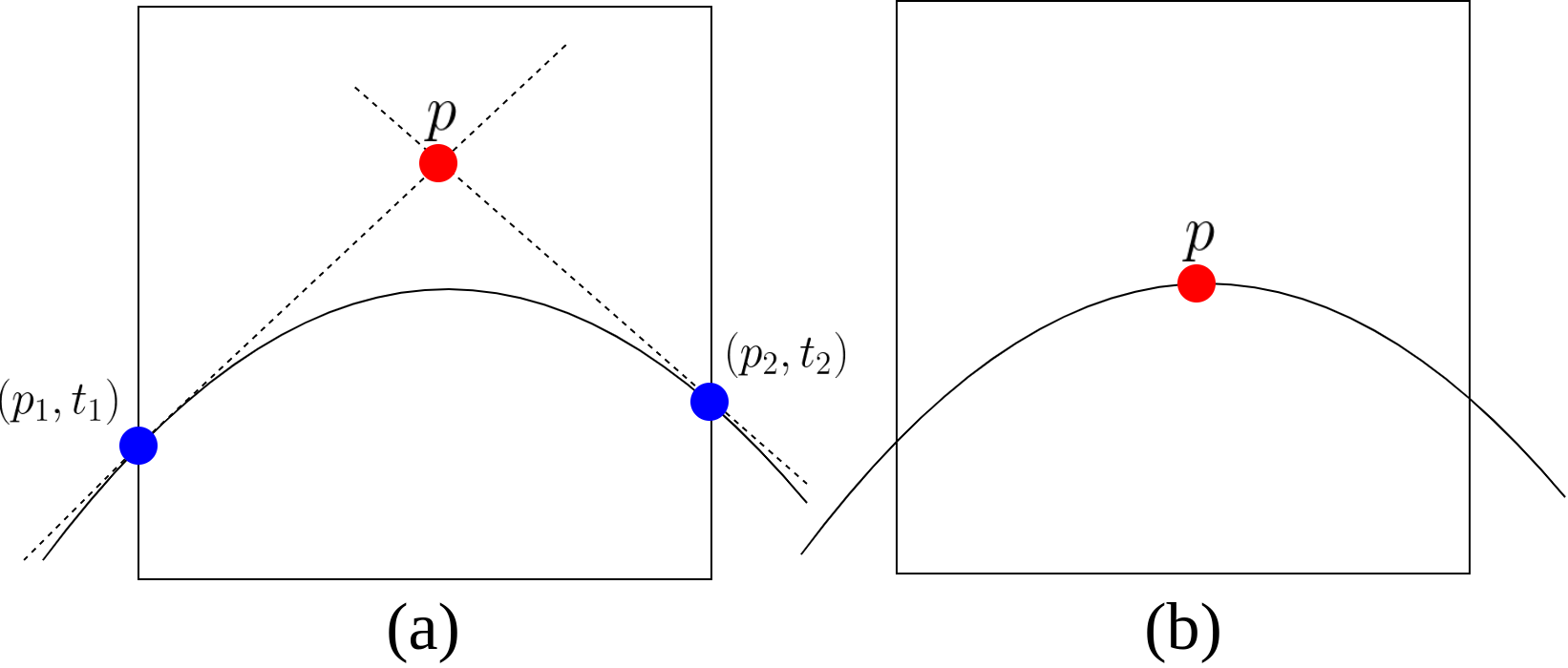}
	\caption{2D illustration for two types of cube point definitions. (a) $p$ is obtained by minimizing a QEF. (b) $p$ is obtained by taking the midpoint of the truncated curve inside the cube.
	}
	\label{fig:cube_point_choice}
	\vspace{-1mm}
\end{figure}

\begin{figure*}[!t]
	\centering
	\includegraphics[width=1.\linewidth]{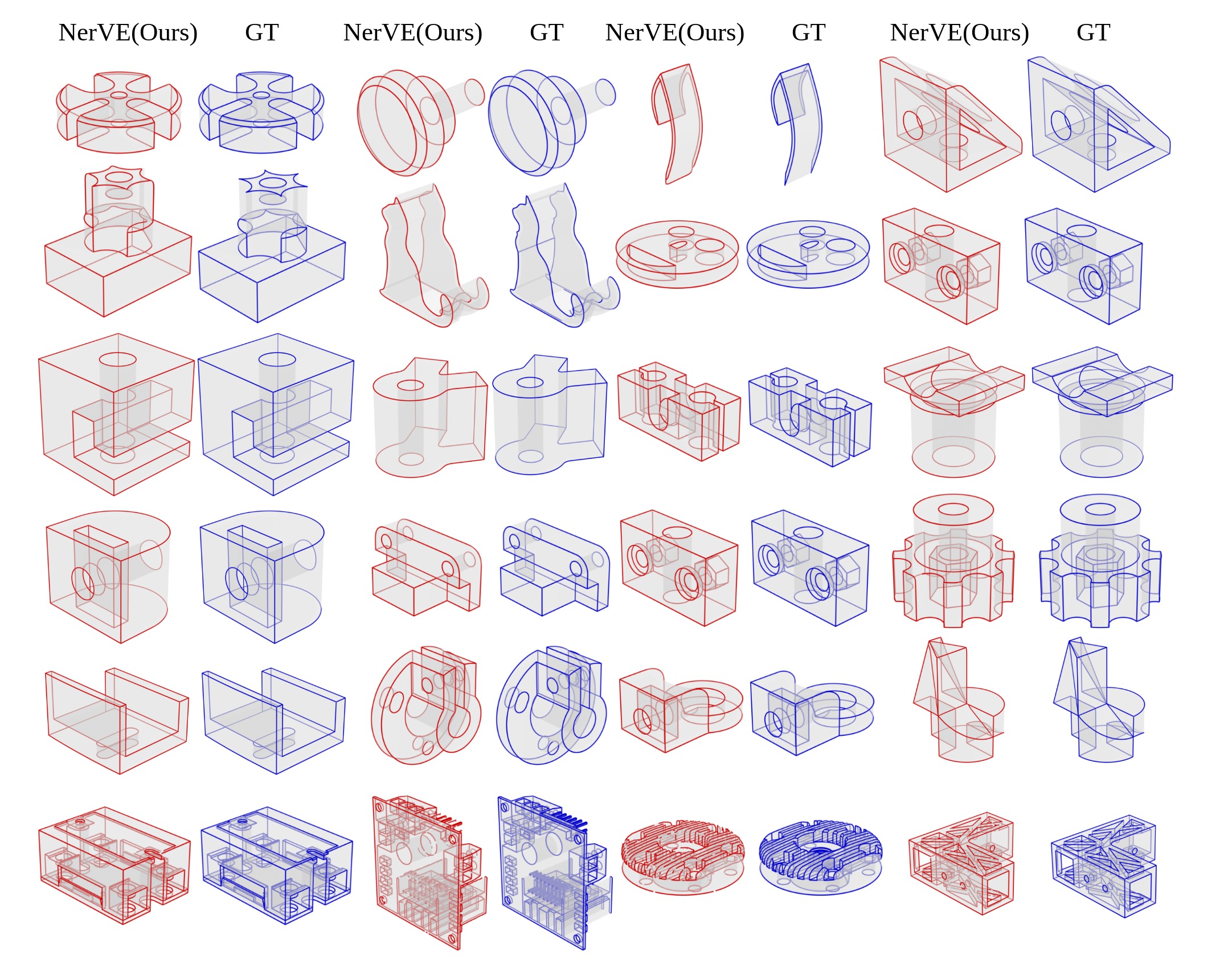}
	\caption{More results of our method. The last row shows the predicted results of our method on 4 complicated cases. }
	\label{fig:more_results}
	\vspace{-4mm}
\end{figure*}

\paragraph{Ablation Study for Cube Point Choice .} We show details for the discussion of cube point choice. In Dual Contouring~\cite{ju2002dual}, the point position in the cube is calculated by minimizing a quadratic error function (QEF) on Hermite data of the surface, which are intersection points of the surface with the cube edges and their corresponding normal vectors. Let $p$ be the point position, it can be calculated by the following minimization:
\begin{align}
    p = \arg \min_x \sum_{i} (n_i \cdot (x - p_i))^2.
\end{align}
where $p_i$ is one of the intersection points with the cube edges and $n_i$ is its normal vector. The surface is approximated as a plane at each intersection point, and the formulation minimizes all the distances from $p$ to all planes in the sense of least squares. One natural counterpart for a 3D curve is to consider the intersection points with cube faces and their tangent vectors. Similarly, we approximate the curve as a line at each intersection point $p_i$ with the cube face, and minimize all distances from $p$ to all lines in the sense of least squares. Let $t_i$ be the tangent vector of $p_i$ (the direction does not need to be specified), QEF for the curve can be formulated as:
\begin{align}
    \min_{x, \forall \alpha_i} \sum_{i} \left \| x - p_i - \alpha_i t_i\right \|_2^2 + \lambda \sum_{i} \alpha_i^2.
\end{align}
where $\alpha_i$ is proposed to enable the problem to be solved by a linear system and $\lambda$ is a weight to balance the two terms. Notice that we only need to solve for $x$, and the system can be easily reformed as a linear system of order 3 to solve for $x$. However, such a type of point position definition does not perform well in the curve restoration as shown in our main paper. Therefore, we finally choose a simple and accurate definition of the point position, which adopts the midpoint of the truncated curve inside the cube. See Fig.~\ref{fig:cube_point_choice} for a visual comparison of these two definitions.

\paragraph{Performance in Higher Resolution.}
To evaluate the performance of \modelName{} in a higher resolution, we perform experiments using resolution $128^3$ and compare the results with lower resolutions, as shown in Table~\ref{tab:performance_reso128}. Using resolution $128^3$ brings more accurate PWL outputs (CD, HD), with slightly fewer scores of \RecallCube{}, \PrecisionCube{}, and \CorrectFace{}. The average inference times of Ours-$32^3$, Ours-$64^3$, and Ours-$128^3$ are 0.15, 0.21, and 0.57 seconds, respectively. On the whole, the performance of \modelName{} can scale well with the increase of voxel grid resolutions. 

\begin{table}[htb]
    \centering
    \resizebox{0.91\columnwidth}{!}{
        \begin{tabular}{l|c|c|c|c|c|c}
        \toprule
                   & \RecallCube{}$\uparrow$ & \PrecisionCube{}$\uparrow$ & \CorrectFace{}$\uparrow$ & \DistancePoint{}$\downarrow$ & CD$\downarrow$ & HD$\downarrow$ \\
             \hline
            Reso $32^3$ & \textbf{0.965}  & \textbf{0.965}  & \textbf{0.940}  & 0.003  & 0.0012  & 0.0714 \\
            \hline
            Reso $64^3$ & 0.958  & 0.960  & \textbf{0.940}  & 0.001  & 0.0009  & 0.0523 \\
            \hline
            Reso $128^3$ & 0.945  & 0.947  & 0.914  & \textbf{0.0005}  & \textbf{0.0008}  & \textbf{0.0484} \\
        \bottomrule
        \end{tabular}
    }
    \caption{\modelName{} performance in different resolutions.}
    \label{tab:performance_reso128}
    \vspace{-4mm}
\end{table}

\paragraph{Ablation on Choice of $k$ in Point Encoder.}
We choose $k=8$ empirically based on the experiments of resolution $32^3$. For the higher resolution $64^3$, using a larger $k$ may improve the accuracy but incur much more calculation cost, as shown in Table~\ref{tab:supp_ablation_KNN}. It is a trade-off of accuracy and efficiency to apply $k=8$ in our experiments.

\begin{table}[htb]
    \centering
    \resizebox{\columnwidth}{!}{
        \begin{tabular}{l|c|c|c|c|c|c}
        \toprule
              & $32^3,k=4$ & $32^3,k=8$ & $32^3,k=16$ & $64^3,k=4$ & $64^3,k=8$  & $64^3,k=16$ \\
            \hline
            CD$\downarrow$ & 0.0017 & 0.0012 & 0.0012 & 0.0015 & 0.0009 & 0.0008 \\
            \hline
            HD$\downarrow$ & 0.0833 & 0.0714 & 0.0669 & 0.0680 & 0.0523 & 0.0491 \\
        \bottomrule
        \end{tabular}
    }
    \caption{Ablation study on choices of k in the point encoder. }
    \label{tab:supp_ablation_KNN}
    \vspace{-5mm}
\end{table}

\paragraph{Ablation on Using PointNet++ or 3DCNN Features.}
We conducted an ablation study with or without using PointNet++ and 3DCNN features. The results of the network predictions and the PWL curves are reported in Table~\ref{tab:supp_PN_CNN}. As shown, both the PointNet++ and 3DCNN features can promote the performance of \modelName{}.

\begin{table}[htb]
    \centering
    \resizebox{1.\columnwidth}{!}{
        \begin{tabular}{l|c|c|c|c|c|c}
        \toprule
                   & \RecallCube{}$\uparrow$ & \PrecisionCube{}$\uparrow$ & \CorrectFace{}$\uparrow$ & \DistancePoint{}$\downarrow$ & CD$\downarrow$ & HD$\downarrow$ \\
            \hline
            wo PointNet++ & 0.9358 & 0.9494 & 0.8517 & 0.0041 & 0.0025 & 0.0982 \\
            \hline
            wo 3DCNN & 0.9383 & 0.9460 & 0.9127 & 0.0040 & 0.0020 & 0.0942 \\
            \hline
            Ours & \textbf{0.9649} & \textbf{0.9650} & \textbf{0.9437} & \textbf{0.0030} & \textbf{0.0012} & \textbf{0.0714} \\
        \bottomrule
        \end{tabular}
    }
    \caption{Ablation study on using PointNet++ and 3DCNN blocks.}
    \label{tab:supp_PN_CNN}
    \vspace{-4mm}
\end{table}

\paragraph{Ablation Study on Post-processing.}
The quantitative results of parametric curves with/without post-processing are CD: 0.008/0.067, HD: 0.224/0.225. Thus, the post-processing is necessary numerically for better parametric extraction.

\paragraph{Ablation Study on data splits.}
Our experiments are all based on the same random split. Here, we test on other two different random splits with resolution $32^3$, the CD results are 0.0013 and 0.0016 (the number in our main paper is 0.0012). As shown, the differences are insignificant.

\paragraph{More Results.} More results of our method are shown in Fig.~\ref{fig:more_results}. Our method can produce reasonable results even for complicated cases, as shown in the last row of Fig.~\ref{fig:more_results}. 

\end{document}